# Symbiotic System of Systems Design for Safe and Resilient Autonomous Robotics in Offshore Wind Farms

**Daniel Mitchell[1], Jamie Blanche[1], Osama Zaki[1], Joshua Roe[1], Leo Kong[1], Samuel Harper[1], Valentin Robu[1,2,3], Theodore Lim[1] and David Flynn[1]**

[1]Smart Systems Group, Institute of Sensors, Signals and Systems, School of Engineering and Physical Sciences, Heriot-Watt University, Edinburgh, EH14 4AS, U.K.
[2]CWI, Centre for Mathematics and Computer Science, Intelligent and Autonomous Systems Group, 1098 XG Amsterdam, The Netherlands.
[3]Algorithmics Group, Faculty of Electrical Engineering, Mathematics and Computer Science (EEMCS), Delft University of Technology (TU Delft), 2628 XE Delft, The Netherlands.

Corresponding author: Daniel Mitchell (e-mail: dm68@hw.ac.uk)

**ABSTRACT**

To reduce Operation and Maintenance (O&M) costs on offshore wind farms, wherein 80% of the O&M cost relates to deploying personnel, the offshore wind sector looks to Robotics and Artificial Intelligence (RAI) for solutions. Barriers to Beyond Visual Line of Sight (BVLOS) robotics include operational safety compliance and resilience, inhibiting the commercialization of autonomous services offshore. To address safety and resilience challenges we propose a Symbiotic System Of Systems Approach (SSOSA), reflecting the lifecycle learning and co-evolution with knowledge sharing for mutual gain of robotic platforms and remote human operators. Our novel methodology enables the run-time verification of safety, reliability and resilience during autonomous missions. To achieve this, a Symbiotic Digital Architecture (SDA) was developed to synchronize digital models of the robot, environment, infrastructure, and integrate front-end analytics and bidirectional communication for autonomous adaptive mission planning and situation reporting to a remote operator. A reliability ontology for the deployed robot, based on our holistic hierarchical-relational model, supports computationally efficient platform data analysis. We demonstrate an asset inspection mission within a confined space through Cooperative, Collaborative and Corroborative ($C^3$) governance (internal and external symbiosis) via decision-making processes and the associated structures. We create a hyper enabled human interaction capability to analyze the mission status, diagnostics of critical sub-systems within the robot to provide automatic updates to our AI-driven run-time reliability ontology. This enables faults to be translated into failure modes for decision-making during the mission. Our results demonstrate that the symbiotic system of systems methodology provides enhanced run-time operational resilience and safety compliance to BVLOS autonomous missions. The SSOSA is highly transferable to other mission scenarios and technologies, providing a pathway to implementing scalable and diverse autonomous services.

**INDEX TERMS** Artificial Intelligence, Autonomous Systems, Digital Twin, Non-Destructive Evaluation, Resilient Robotics, Safety Compliance, Sensors, Symbiotic Systems and System Ontology

## I. INTRODUCTION

Offshore wind farms are large complex structures where operations, maintenance and servicing pose significant technological and economic challenges [1]. Typically, 80% of the cost of offshore Operation and Maintenance (O&M) is attributed to transporting engineers to remote sites for asset inspection [2]. In 2018, the Crown Estate held offshore leasing rounds for Scotland, England and Wales, representing a combined UK increase of 74% for offshore wind installations. Furthermore, the UK government has committed to produce 40GW of offshore wind, with 1GW utilizing floating foundations by 2030 [3]. The Global Wind Energy Council estimates the UK will represent 34% of European offshore wind capacity, which is projected to reach 89GW by 2030 [4]–[6].

Advancements in fault detection methods utilizing novel sensing technologies, data analysis and modelling, have reduced both fatal incidents and the need for logistically expensive and time-consuming human interventions [7], [8]. However, as demand for offshore wind energy grows alongside the political will for a sustainable energy driven, post-COVID-19 economy, the next generation of offshore wind farms will require significantly larger constructs and data-driven systems. With wind farms planned further from shore, residential robots with cognizant autonomy will be an increasingly required feature. Another challenge to the offshore wind energy sector is its remote and hazardous environment, resulting in inclusion within the high-risk register. In 2016, 344 high potential incidents in the global offshore wind sector were reported. This contrasts with the observed 2019 data, with 252 incidents of the same severity [9]. This reduction may be attributed to the implementation of higher safety standards and improved reporting and transparency. With wind farms planned further from shore, emergency response times to severe incidents pose serious implications.

Robotics and AI (RAI) is a promising field of significant innovation aligned with many of the safety, recruitment, operational and planning challenges in offshore wind O&M [10]. Most offshore wind operators have included RAI within their commercialization roadmaps for operational and end-of-life services [11]–[16], where improved safety results from the removal of personnel from dangerous environments, the reduction of asset downtime and reductions in O&M costs. However, international safety regulators, such as the civil aviation authorities, have identified that run-time safety and reliability of autonomous systems are key obstacles in Beyond Visual Line of Sight (BVLOS) missions for unmanned aerial vehicles in twelve European countries [17]. We also highlight dynamic conditions, such as weather conditions, smoke, steam and mist, as challenges which reduce the reliability of sensors onboard robotic platforms [8], [18], [19]. Despite advancements in RAI, there are several barriers that limit expansion in the offshore domain, which include technological, regulatory and commercial challenges. While much RAI research focuses on convergence to enhance autonomy through learning, RAI systems cannot deal with situations where there is an absence of data [20], [21]. These points highlight the need of symbiotic digitalization, not only across O&M, but across the entire lifecycle management of the offshore wind farm.

Symbiotic RAI relationships consist of several elements (robotic platforms, humans and smart environments) that could cooperate when performing tasks [22]. Three basic types of symbiosis exist: mutualism, commensalism and parasitism [23], [24].

State-of-the-art symbiotic robotic systems generally focus on the singular concept of cooperation, collaboration or corroboration between robotic platforms. We enhance the symbiosis between systems by including Cooperation, Collaboration and Corroboration ($C^3$) between robotic platforms based on the concept of $C^3$ governance. All three relationships are based on internal and external (inter-intra) objectives and rules, such as a predefined mission. By considering symbiotic relationships, in terms of $C^3$, to execute functional, operational, planning, and safety activities, a future capability to systematically characterize trustworthy relationships is facilitated.

As in nature, data transactions and system awareness are governed by communication rules. This paper introduces and proposes a Symbiotic System of Systems Approach (SSOSA) and Symbiotic Digital Architecture (SDA) using a top-down assessment of RAI and O&M challenges. This creates a symbiotic digital framework that includes functional, operational, planning and safety requirements of resilient autonomous missions, resulting in a new hyper-enabled environment for knowledge sharing, operational and safety requirements. In this instance, we define resilience as the capability to adapt and survive in an autonomous mission in response to internal and external variables, such as reliability. We evaluate resilience via mission success in compliance with live safety cases and system reliability variables.

This paper provides an example showcasing symbiotic collaboration across different systems utilizing a commercial off-the-shelf robotic platform conducting an autonomous confined space asset integrity inspection, where proof-of-concept autonomous mission evaluation videos can be accessed via Mitchell *et al.* [25]–[27]. Symbiotic $C^3$ governance is achieved using a run-time reliability ontology on the inspection robot together with distributed edge analytics to improve holistic systems visibility in near-to-real-time. This provides a continuous strategic view of the asset but never at the expense of safety governance.

We define two paradigms in advancing our roadmap to trusted autonomy and self-certification. These paradigms represent progressive levels of safety compliance and reliability leading to advances in successful servitization to meet the requirements imposed by an increasingly automated offshore environment. This paper focuses on tier 1 - '*Adapt and Survive*' with the intention to serve as a research direction into tier 2 - '*Adapt and Thrive*'.

*Tier 1 - Adapt and Survive* - Where an autonomous mission or service has predefined mission objectives. The system can evaluate: the implications of a scenario of variables from the environment, infrastructure, human interaction and robot reliability; sharing knowledge with and collaborating with a remote human observer; mitigating known and unknown threats to the resilience and safety case of the autonomous mission. Survivability featuring mutualism and commensalism and the completion of a mission without violating safety governance or mission objectives.

*Tier 2 - Adapt and Thrive* – Enhancing the capabilities of *Tier 1* through a recommender system for multi-objectives missions driven by a knowledge distribution map for the human observer. This includes the platform assessing unforeseen circumstances, their consequences and setting suggestions for mission optimization, further developing a symbiotic relationship where cyber physical systems are deployed. These capabilities can feature aspects of parasitism to ensure a platform can thrive but not at the expense of another platform.

While the primary application of the SDA, developed from our SSOSA, is the offshore renewable energy sector, the intention is the wider application to operational and resilience requirements for resident and BVLOS autonomous systems. The digital environment provides a means of creating new information streams on critical front-end systems and provides an operational decision support system with full, bidirectional, interaction between the robot and remote human observer.

This paper is structured as follows: Section II reviews the state-of-the-art in RAI within the sector context of offshore wind, where the emphasis is towards autonomous systems, asset integrity sensing and management, reliability and human-robot interactions. Section III describes our methodology. Symbiotic interactions and $C^3$ governance are discussed in Section IIIA, where we outline the barriers of current symbiotic systems and discuss types of interactions which exist in symbiosis against the state-of-the-art. Section IIIB focuses on safety compliance and resilience in autonomous systems via the introduction of our SSOSA, which includes our system integration process and SDA design, encompassing our Tier 1 *'Adapt and Survive'* paradigm. Section IV outlines our symbiotic implementation and its digitalization exemplified through an autonomous confined space asset integrity inspection mission evaluation. Section V presents millimeter-wave sensing and its potential to enhance symbiotic asset integrity management. Section VI concludes by summarizing the framework for symbiotic RAI, the proposition of *'Adapt and Survive'* theory building and future steps to transition to Tier 2- *'Adapt and Thrive'*. To aid reader understanding of the paper structure, Figure 1 illustrates the motivations and impacts of each subsection of this manuscript.

## II. LITERATURE REVIEW

Offshore wind turbines present unique engineering challenges due to operation in the harsh offshore environment, with correspondingly higher failure rates compared to their land-based counterparts [28]. Wear-out failures and random system faults pose the greatest challenge to productivity. These faults are difficult to monitor and can result in permanent damage, in addition to causing other subsystems to fail [29], resulting in both operational and economic costs. A study of offshore wind turbine failure modes identified the pitch control and hydraulic mechanisms, generator, gearbox and blades as the largest individual system failures at ~13%, 12.1%, 7.6% and 6.2% respectively [28]. While several surveys on wind turbine condition monitoring and fault diagnostics exist, research on symbiotic Robotics and Autonomous Systems (RAS) for offshore wind energy O&M purposes to meet the Structural Health Monitoring (SHM) and Condition Monitoring (CM) challenge is limited.

In this section, we provide a top-down approach via a review of the state-of-the-art in RAS/I, which we evaluate

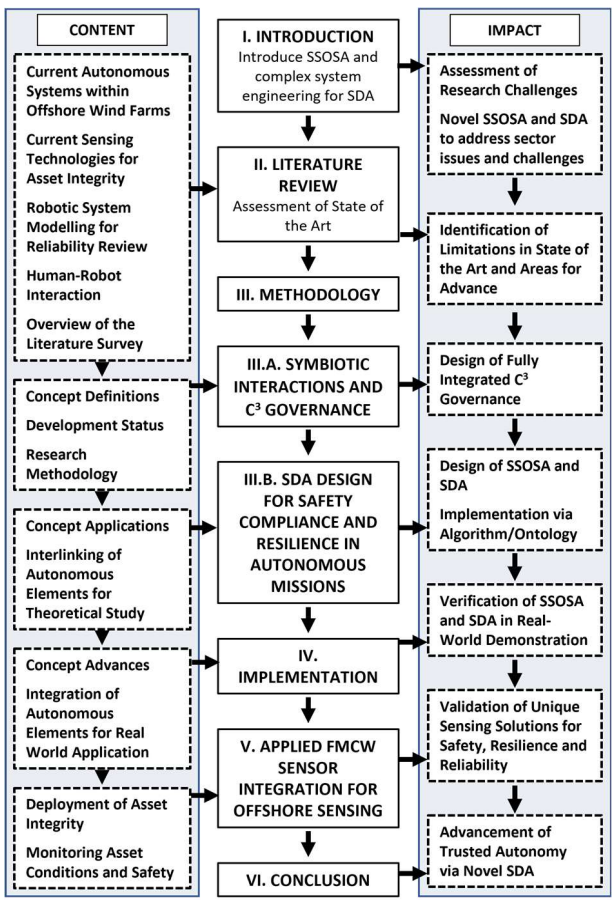

**Figure 1** Paper structure detailing content and impact for each section.

against the three key capability criteria of the *'Adapt and Survive'* paradigm:
1) The ability of the field deployed robots to self-certify their state of health during mission run-time,
2) The ability of a human to interact collaboratively with the robot with an enhanced hyper-enabled situational report,
3) The ability to assure safety compliance and resilience of the mission within a dynamic environment, which can consist of system, environment and operational unknowns.

Based on these three capability challenges for offshore robotics: Section IIA summarizes state-of-the-art autonomous robotic platforms being used in offshore wind farms and their environmental analogues. Section IIB evaluates sensing technologies and requirements for offshore asset inspection while Section IIC reviews system modelling for robotic platforms to work safely and reliably within the infrastructure and environment. Section IID provides an overview of Human Robot Interaction (HRI) and the development of Digital Twins. Section IIE presents an overview and critical summary of the literature survey.

### A. Survey of Autonomous Systems for Offshore Wind Farm Deployment

Most robotic systems used to support offshore wind O&M functions are deployed service robots designed primarily for logistical applications in non-manufacturing environments. Robots used in an offshore role can be classified as follows:

- Autonomous Underwater Vehicles (AUV) deployed to inspect foundations and underwater cabling.
- Unmanned Aerial Vehicles (UAV) used for inspections of wind turbine blades.
- Autonomous Surface Vessel (ASV) enabling autonomous cargo transfer via handling systems, logistics management and system analytics.
- Crawler robots to inspect wind turbine exteriors such as tower and blades.
- Autonomous Ground Vehicles (AGV) deployed within substations and onshore operations.
- Robot railed systems employed in substations and nacelles utilizing infra-red and other cameras.

For inspection, RAS must provide condition monitoring and fault diagnostics to meet a pre-scheduled inspection regime. Commercial services and ongoing research predominately focus on robot-based approaches to inspect the external structure of the turbine. The current methods used to inspect these high-value assets represent time consuming and dangerous work for rope access crews working at height and in highly changeable offshore conditions [30]. Thus, a key metric in the future of robot development in the offshore sector relates to how robotic operations in this remote environment can be made safer for the robotic system, human operators and the offshore asset, while simultaneously maintaining a trustworthy BVLOS capability. The following subsections provide a critical analysis of these robotic systems via comparison to the needs imposed by persistent trusted autonomy: safety, resilience and reliability.

#### 1) WIRE-DRIVEN ROBOTICS
Wire driven robotic systems for wind turbine inspection and maintenance represent automated rope access systems, offering an inspection solution by saving time, cost and labor, while also ensuring safety. These systems have many designs; however, the consistent trend is an open frame device housing sensors or repair equipment that is maneuvered along the wind turbine blade or tower. Sensor configurations include ultrasonic, infrared thermographs and visual spectrum high resolution imagery for inspection of bonded spar joints, leading edges and trailing edges. One major limiting factor of these devices is the requirement for physical contact with the blade or tower, leading to the potential for damage to the wind turbine structure. Robotic systems deployed via wire-driven systems include water jet and brush-based cleaning systems, with an onboard supply of cleansing [31], [32]. An alternative to wire-driven devices is the railed robotic system, which uses the blade leading edge as a guiding rail, and utilizes an on-board winch and camera system for blade leading edge inspection and repair [33].

#### 2) CLIMBING ROBOTS
The development of climbing robots focuses on the use of suction for adhesion, where a vacuum is created within the surface in contact with the blade, resulting in sufficient grip for the robot to climb the wind turbine tower or blade. A small, remotely controlled tracked vehicle was developed by General Electric with an on-board camera for visual inspection of blades [34]. A significantly larger crawling ring robot has been designed by researchers at London South Bank University and is capable of internal wind turbine blade inspection and is based on axial X-ray tomography [35], [36] Climbing robotic systems are limited by the need for physical contact, in addition to slow operation time.

#### 3) UNMANNED AERIAL VEHICLES (UAV)
Aerial inspection with drones offer high mobility and diversified sensing capability, however, they are limited by payload, flight envelope and mission endurance [37]. The latest advancements include multi-directional aerial platforms that can fly in almost any orientation and direction [38]. Used as an inspection RAS platform, multirotor drones can perform complex sensing and manipulation tasks. As a result, UAVs represent a maturing method of asset integrity, with significant uptake from the offshore and onshore energy sectors for pipeline, platform, gas flare and power line network inspection, in addition to search and rescue roles [38]–[40].

### 4) COMBINED-ROBOT PLATFORMS

Developments in the versatility of robotic systems are driven by the need for potentially resident systems to adapt to operational needs. An example of this is the development of a multi-copter with visual and Light Detection and Ranging (LiDAR) sensors for asset inspection [41]. The operational use was to provide reconnaissance and to guide slower climbing robots to analyze structural components, conduct localized non-destructive inspection and to repair small defects.

Development of a prototype friction-based climbing ring robot (SMART— Scanning, Monitoring, Analyzing, Repair and Transportation), coupled to a large industrial manipulator for higher payloads, represents another example of a combined robotic system. However, significant limitations within this system are the very large size of the robot, requiring safe and secure access to the base of the tower and rendering it difficult to deploy in challenging terrain on land and in the offshore environment [42].

The MEDUSA project represents another example of a multi-robot platform for autonomous operation and maintenance of offshore wind farms [43]. This aerial-aquatic RAS combination offers the unique ability of operation in both aerial and marine environments. However, this configuration, where the payload of the UAV is the AUV, results in a severely limited operational envelope, rendering field deployment at this scale impractical.

### 5) UNDERWATER ROBOTICS

Robotic operations involving water have tended to rely on personnel and the complex integration of a range of expensive vessels and vehicles. However, the development of independently operating subsea robotic platforms are now becoming commonplace and a mainstay of the offshore industry for site surveys and inspections of infrastructure [44].

Hugin Endurance is an AUV capable of completing situational awareness scanning, mapping and inspection operations for durations up to 15 days. Such AUV systems can be equipped with a range of sensors including HiSAS synthetic aperture sonar, a wide swath multibeam echo sounder, sub-bottom profilers and magnetometers, in addition to current and turbidity sensors [45]–[49]. AUVs are typically used within offshore wind farms in environmental surveys and wind farm planning phases.

The REMORA project represents collaborative underwater robots designed for inspection and repair of underwater foundations and rigs [50]. The modular design of these robots result in a collaboration, wherein other REMORA robots are able to merge into a larger device through a connection mechanism to overcome environmental challenges that would overwhelm single platforms [51]. This unique function results in increased resilience via shared system awareness and fault recovery.

To better manage dynamic situations for underwater missions, Carreno *et al.* proposed a Decentralized Heterogeneous Robot Task Allocator (DHRTA) algorithm [47], [52]. While not specific to any robotic platform, its purpose is to improve the task planning of offshore underwater missions for AUVs.

Table I provides a comparative analysis of deployed robotics and enabling technologies against our criteria for self-certification, human-robot interaction and safety compliance. An immediate observation pertains to the identification of knowledge gaps in run-time safety, reliability and resilience. Our top-down analysis identifies that, while systems for deployed integrity inspection are reaching more sophisticated levels of development, no systems are currently able to self-certify their state of health in run-time and none of the systems are used in a resident capacity offshore. Current collaborative robotic interactions are low level and tend to not feed their data to a digital twin of the asset network for human/robot collaboration or intervention. Many of the systems reviewed require human operation and supervision, are not self-deploying and are most often singularly deployed robotic systems that are not designed for, or capable of, multi-robot interaction and collaboration. Furthermore, in most cases, the robotic systems evaluated in this section require significant support to operate in the offshore environment, resulting in continued human presence in hazardous areas. These observations reinforce the need to develop symbiosis with other robotic systems to meet the needs of the dynamic and multi-role requirements to be found in the offshore environment for safe BVLOS operations.

### B. Sensing Technologies for Asset Integrity

Sensing technologies are integral to inspection and mission guidance for deployed robotics. The current state-of-the-art in wind turbine asset inspection is centered on UAV platform-based acquisition of high-resolution images (visual and infrared) for expert analysis and to identify/infer regions of notable damage [8].

State-of-the-art embedded SHM strategies require sensor integration during offshore infrastructure and wind turbine systems design and construction phases. The embedded SHM paradigm represents a more recent strategy due to the ubiquity of wireless telemetry, however, older offshore wind infrastructures require retrofitting of CM and SHM sensors, adding to operational and capital expenditure demands of aging wind turbines structures. In older wind turbine designs, the practice of overengineering has compensated for the potential for undetected operationally induced damage in the offshore environment, where inspection is less frequent and costly. Consequently, overengineering prevents the premature decommissioning of the wind turbine asset relative to the expected remaining useful life [53]. However, as wind turbine designs become ever larger, the practical margins for overengineering become smaller due to the cost and weight of additional structural material, necessitating the integration of SHM and CM systems from the design inception stage.

TABLE I
SUMMARY OF CAPABILITIES OF ROBOTIC PLATFORMS AVAILABLE FOR OFFSHORE WINDFARMS AND ANALOGUES

| Robot Platform | Task/ Function | Year | Resident System | Onboard Reliability Analysis | Multi-Robot Collaboration | Human-Robot Interaction | Ref. Works |
|---|---|---|---|---|---|---|---|
| **Wired/ Crawling/ Railed** | Inspection | 2008 | ✗ | ✗ | ✗ | 2D Graphical User Interface (GUI) | [36] |
| | | 2010 | ✗ | ✗ | ✗ | 2D GUI | [31] |
| | | 2012 | ✗ | ✗ | ✗ | 2D GUI | [34] |
| | | 2016 | ✗ | ✗ | ✗ | 2D GUI | [35] |
| | Maintenance | 2012 | ✗ | ✗ | ✗ | Micro-controller Interface | [32] |
| | | 2017 | ✗ | ✗ | ✗ | VR live feed | [33] |
| **UAV** | Inspection | 2020 | ✗ | ✗ | ✗ | Betaflight 4.1 firmware | [38] |
| | | 2020 | ✗ | ✗ | ✗ | Flight control firmware | [39] |
| | | 2021 | ✗ | ✗ | ✗ | Flight control firmware | [40] |
| **AUV** | Inspection Maintenance | 2017 | ✗ | ✗ | ✗ | 2D GUI | [50], [51] |
| | | 2020 | ✗ | ✗ | ✗ | 2D GUI | [46], [48], [49], [111] |
| | | 2020 | ✗ | Limited to onboard capability analysis | ✗ | ROSPlan Interface | [47], [52] |
| | Offshore Geosensing | 2014 | ✗ | ✗ | ✗ | 2D GUI | [44] |
| **AGV** | Inspection | 2021 | ✗ | ✗ | ✗ | 2D GUI | [112]–[115] |
| | | 2021 | ✗ | ✗ | ✗ | 2D GUI | [116] |
| **Multi-robot** | Inspection Maintenance | 2020 | ✗ | ✗ | ✓ | Multi-control system via 2D GUI | [43] |
| | | 2020 | ✗ | ✗ | ✓ | Multi-control system via 2D GUI | [41] |

A key issue in the operation of wind turbine blades is the presence of multiple defects resulting from errors in manufacture, installation or subsequent operation of the blade. While wind turbine blades have significantly increased in size, the manufacturing process has remained largely the same [54]. As wind turbine blades have evolved into larger structures, towards 100 meters in length, and are exposed to greater forces, the need to inspect for operationally induced defects has become a regular requirement to be performed, often bi-annual, to uncover structural defects that may require intervention, repair or replacement [55], [56]. Common defects which can be classified into distinct types within wind turbine blade structures include porosity due to delamination, water ingress, leading edge erosion, stress induced cracking and lightning strikes [8]. Detecting and categorizing defects remains key for asset integrity, where multiple sensing methods have been employed to classify defect types and to quantify the severity of a detected defect.

The most common method for wind turbine inspection is visual spectrum photography and photogrammetry devices. This method is effective at identifying regions of impact damage or erosion on the blade surface and can identify cracks and distortions that are evident from the surface of the blade. Limitations of this method are related to lighting and shadows cast at differing times of the day, presenting issues with analysis algorithms [56]–[58].

Infrared thermography is a non-contact method to inspect the thermal conductivity of a target, where observed contrasts indicate latent defects within the structure, where interruptions in the material continuity of the structure result in contrasts in thermal radiation from the target surface. A limitation of this method relates to defect depth and defect radius, where beyond a certain depth within a structure, certain defects are undetectable. This inspection method is also vulnerable to surface irregularities and environmental or atmospheric temperature variations [59], [60].

Ultrasonic, or pulse echo measurements, have been successfully applied to the detection of cracks and delamination defects, with additional measurands simultaneously recorded, such as material thickness and defect position or orientation. A key limitation of this inspection method is the need for very close proximity or physical contact with the blade structure, often requiring a couplant [61]–[63]. The state-of-the-art in acoustic sensing is represented by the Electromagnetic Acoustic Transducer (EMAT), which provides contact ultrasonic inspection. Applications include inspection of curved or flat metallic surfaces, splash zone inspection, subsea pipeline inspection and wind turbine jacket foundation and tower inspection. However, the system is limited to metallic surfaces and requires direct contact with the material under inspection, albeit without a couplant [64], [65].

Microwave Frequency Modulated Continuous Wave (FMCW) radar is an emergent technology for condition monitoring of multiple types of materials. The non-contact sensing mechanism enables surface and subsurface detection of faults within porous (low dielectric) structures, where defects can be detected deep within a wind turbine structure. Applications of FMCW in the K-band (18 – 26 GHz) have included the detection of fluid ingress and delamination features within wind turbine structures; common defect types in manufacturing and operationally induced defects [8] [18]. Microwave radar in the W-band (75-110 GHz) has also been successfully tested as an embedded sensor on wind turbine towers, allowing for the analysis of vibration in major wind turbine structural components to inform asset prognostics [66], [67]. Continued development of super high frequency, extremely high frequency and terahertz devices, is an area of frontier sensor research for all asset integrity sectors and seeks to exploit the unique properties of novel materials [68].

Table II details the key candidate technologies, in addition to their measurands and limitations, for embedded and robotically deployed sensing solutions for run-time capable wireless telemetry of data and application to our SDA. Table II indicates advancements in robotic deployment will result in sensor suites that require less power, are lighter in weight and will result in robotic platforms such as UAVs, AGVs and crawlers that are able to easily integrate the sensors as payloads for asset integrity inspection. A key observation from this literature survey is that, while many SHM/CM methods have been applied to deployable systems to date, there remains a lack of robust sensing methods capable of *in situ* measurement via structural embedding and wireless telemetry [69]. These qualities are essential pre-requisites for Internet of Things (IoT) and Digital Twin (DT) applications and will be crucial for the integration of real-time data with the synthetic digital environment. This review of deployable sensing also identifies that few robotic systems are capable of a wide array of sensing modalities to provide full-field measurements, using the multiple sensing methods required to interrogate the metal and composite structures of offshore renewable infrastructures. Consequently, we identify that there exists a need for resident autonomous systems to deploy low power, highly tunable and lightweight sensor suites for run-time composite structural analysis. As a candidate technology for corrosion on metal surfaces and internal defect detection in composites, we identify millimeter-wave radar as the sensor mode with the greatest potential for robotic deployment for offshore wind turbine integrity analysis. Consequently, millimeter-wave sensing is utilized during this autonomous mission evaluation.

### C. Robotic System Modelling for Reliability

A primary motivation for the use of reliability ontologies within robotics is the application of knowledge-based approaches, offering an expandable and adaptable framework for capturing the semantic features to model robot cognitive capabilities. This results in an agile and rapidly tunable capability of dynamic safety compliance and mission operability requirements. These capabilities directly impact the real-time safety case, reliability and resilience of a robotic system. The developed ontology can be applied to several tasks that humans and robots can perform cooperatively within a defined infrastructure, mission profile and environment [70].

Though progress has been made to create Core Ontology for Robotics and Automation (CORA), developed in the context of the IEEE Ontologies for Robotics and Automation (ORA) Working Group, creating a complete framework is a highly complex task [71]. The IEEE Standard ORA [72] describes an overall ontology as including key terms and their definitions, attributes, constraints and relationships. Sub-parts of this standard include a linguistic framework, generic concepts (an upper ontology), a methodology to add new concepts, and sub-domain ontologies. The resulting core ontology described in [72] was utilized in projects such as [73], [74].

In our review, only ontology frameworks relating to the ability to model the reliability of a system were selected. Knowledge Processing for Robots (KNOWROB) is widely used, and arguably one of the most influential projects, due to the use of knowledge-based reasoning for autonomous robots. The ROSETTA ontology focuses on the relationships between robotic devices and skills. Semantic Web for Robots is an ontology implemented by [75] for robotic components and uses a collaborative knowledge acquisition system for encoding information about robotic devices. In the PANDORA framework [76], ontologies are used as a means for the robot to organize knowledge about the world, not just in geometric concepts, but by attaching a semantic label. The project aimed to demonstrate the challenges of integrating autonomous inspection of an underwater structure, autonomous location, cleaning and inspection of an anchor chain and autonomous manipulation of a valve from an undocked vehicle. The PANDORA framework also investigated the relation between action capabilities and the planning system.

TABLE II
STATE-OF-THE-ART SENSING METHODS FOR WIND TURBINE INSPECTION.

| Sensor Type | Application | Wind Turbine Structural Element | Defect Type | Measurand | Limitations | Ref. Works |
|---|---|---|---|---|---|---|
| **Visual Camera** | Surface defect detection | Tower, Nacelle and Blade | Cracks Material Loss Surface Corrosion | Visible spectrum contrasts | Shadows (time of day) Surface observations only | [55]–[58] |
| **Infrared** | Near subsurface defect detection | Tower, Nacelle and Blade | Delamination, Water ingress Concealed inclusions Material loss | Thermal contrasts caused by changes in thermal conductivity or continuity of material | Affected by ambient conditions Only able to infer presence of near to surface defects | [59], [60] |
| **Vibration Accelerometry** | Structural or Component defect detection | Tower, Nacelle and Blade | Structural degradation Effects of wear on moving parts | Vibrational analysis of embedded accelerometer data | Requires embedded sensing at construction phase or retrospective installation | [117]–[121] |
| **Electrochemical Sensing** | Corrosion of Metal Structures | Tower, Nacelle | Structural degradation due to corrosion Material loss | Variation in open circuit potential | | [122], [123] |
| **Ultrasonic/ Acoustic** | Active Acoustic Internal Feature Detection | Tower, Nacelle and Blade | Delamination, Water ingress Concealed inclusions Material loss Corrosion | Material density/impedance contrasts | Requires physical or very close contact with asset May require couplant | [40], [61]–[65], [124] |
| **X-Ray** | Internal Feature Detection | Blade | Delamination, Water ingress Concealed inclusions Material loss | X-ray impedance contrasts | Power draw and weight require large structural anchor and cross balancing at wind turbine tower | [36] |
| **Millimeter-wave radar** | Vibration detection Surface defect detection Internal defect detection (low dielectric) | Tower, Nacelle and Blade | Delamination, Water ingress Concealed inclusions Material loss | Return signal amplitude response variations in reflection coefficient at each material boundary | Cannot detect beneath metal layers | [8], [18], [66]–[68] |

These frameworks have not addressed the challenges relating to diagnostics and prognostics but provide contributions to relevant hardware configurations. In addition, the current state-of-the-art does not provide semantic relationships within their frameworks, except for KNOWROB which uses a single "Depends-on" relationship, resulting in a constrained model [75].

In this section, we identify that for a complex system of systems, more semantic relationships are required to describe the symbiotic relations which must define the relationships between heterogeneous conditions and objects found in a configured environment. For example, the relationships between parts of the infrastructure and RAS, the RAS and the environment via static passive or active sensors, the communication between RAS, and lastly the human interactions with the robots. Therefore, modifications to existing models must be made to develop a system of systems able to meet the capability criteria stated in Section II.

### D. *Human-Robot Interaction (HRI)*

HRI has evolved significantly and, while application dependent, encompasses a wide range of user interface types. Current technologies include: natural language processing via speech recognition, mouse-based interaction via desktop or laptop computing, or gesture interaction via mixed/augmented reality systems [77]. The rapid development of intuitive HRI represents a key element in the efficient and safe bidirectional transfer of knowledge, and intelligent collaboration, cooperation and corroboration, between an operator and their autonomous assets.

Of key importance is the advent of the DT, which serves not only to virtualize but also to act as a data and information transaction hub for HRI, making RAS/I data more intuitive to interact with and navigate.

Continuous tasks are generally less intuitive with voice commands than with orthodox input systems, such as a game controller. The result of this is that normal operations, such as moving a robot manually to a new location, do not represent the best use of this technology. However, utilizing speech recognition to instruct the robot to autonomously move to a pre-determined location (e.g., "transformer" or "base station") allows these operations to rely on voice-based commands. Speech recognition technology enables personnel to interact/work with autonomous systems, represents an example of human-robot cooperation and has been employed in sectors dealing with large amounts of data [78], [79]. Speech recognition also facilitates operation for a wider range of users, with findings that it is second only to touch control for speed of operation for both younger and older users [80], [81]. This technology provides the ability to use DTs on a wider range of devices than the traditional desktop computer, allowing access via alternative devices, such as smart glasses or mixed reality headsets, with no physical input and allows the operator to perform actions quickly and on-site. Similarly, the use of 'call out' instructions can be implemented and allow for devices to be controlled with no physical inputs.

Digital twinning is a developing technology, where design standards are still being created. For example, the Microsoft HoloLens provides basic built-in gestures for use in applications. An 'air tap' gesture and a 'drag' gesture allows for the use of radio buttons and sliders in 3D space, as well as the repositioning of on-screen elements. The novelty provided by mixed reality devices is the ability to scan the surrounding environment, placing the virtual robot in the real area of operations to evaluate the suitability of the space for robot operation. Gestures enhance operator ability, whilst not being detrimental to the performance of the robot [82].

Ensuring that operators can interact with a DT simulation offers benefits to operators as unexpected and undesirable events are reduced [83]. Advances in more widely available computing hardware have enabled widespread use of DTs to monitor and simulate possibilities, outcomes and provide training in various fields across industry [84]–[88].

This subsection has provided an outline of the state-of-the-art in HRI and has described the development of DTs to allow ease of knowledge exchange between the asset and human operator, via RAI systems. A key observation in this section is that HRI remains heavily reliant on 2D Graphical User Interface outputs and either remote human control or basic robotic autonomy via pre-programmed systems. These methods lack the adaptive run-time situational awareness and autonomous decision-making required to operate in a BVLOS or hazardous environment setting. This section further illustrates that the underlying principle of DTs, via HRI methods, is to provide modelling capability in a virtual environment to run simulations of intent before committing to real world actions and their consequences. Building on the state-of-the-art will require advances in graphical symbolism for digital model contextualization, speech recognition and teleoperation [89], [90]. The future of HRI needs to maintain an optimally balanced human-robot autonomous system fleet with full $C^3$ in operation. Achieving this will require scalability, while also supporting operation and planning decisions via synchronization with a DT. Adaptable to the end user needs of many sectors, the design of a scalable symbiotic architecture to create the rules necessary will advance system overview, resilience, reliability and safety. This builds on the necessary safety requirements to advance human trust in autonomous systems, as defined in our three capability criteria, and acts as a base requirement to achieve the transition from Tier 1 "*Adapt and Survive*" to Tier 2 "*Adapt and Thrive*".

### E. Overview of the Literature Survey

This literature review has examined the state-of-the-art of robotics and sensors deployed in an offshore (or offshore analogue) role. The critical observations we make from the development of robotics for the offshore sector are that research to date has focused on engineering for single purpose deployments limiting RAS to individual tasks such as inspection and maintenance. Commercial Off the Shelf (COTS) robots are commonly used within research as they allow for rapid development, however, are severely limited in terms of design scope and inherently unable to unlock the full potential of autonomous services.

In terms of sensing, there is a requirement for RAS to deploy payloads capable of surface and subsurface inspections of metals and composite structures. To advance this, these sensors require modular design for ease of transfer from robot to robot. In addition to, the embedding of in-situ sensing from the design stage to improve holistic overviews.

As offshore wind farms are commissioned further offshore, it becomes more economical and safer for robots to operate resident to the assets they inspect. Reliability modelling will become essential in standardization process of BVLOS operation. The complex system of systems requirements for reliability modelling necessitates development of semantic relationships.

We identify that the developments required for HRI relate to sensor integration, $C^3$ governance and scalable synchronization with DTs in real-time. Where we also identify the need for data telemetry via front line technology capabilities, a DT captures the information required to be shared across the synthetic environment.

This review finds that critical development concerning resilience, reliability and robustness as regulatory elements of safety compliance is limited. To achieve high fidelity operational inspection and maintenance, there is a requirement for multi-system collaboration to overcome the environmental challenges that exist offshore, reinforcing the need for platform agnostic symbiotic systems.

This literature review identifies that to achieve safety compliance and resilience in dynamic environments, the design of scalable symbiotic architectures will be key to unlocking the full potential of autonomous systems. Section III will describe our methodology on the design of symbiotic interactions and how it unlocks fully collaborative robotics via the design of our SDA.

### III. METHODOLOGY

To meet the capability gaps identified in Section II, our strategy is to build upon what we term as "$C^3$ governance", or systems capable of "Collaboration, Cooperation and Corroboration", via design of autonomous systems that are reliable, resilient and safe to operate in proximity to humans and infrastructure during BVLOS operations.

For operational ease, we assume stable communications and charging strategies to focus on demonstrating the application of our SSOSA via our systems engineered SDA. Our research direction is further illustrated in Figure 2, which details our Tier 1, '*Adapt and Survive*' paradigm and shows the transition from local to global hierarchal steps in the symbiotic digital ecosystem. This figure illustrates the self-organization of individual subsystems and the resultant emergent symbiotic behaviors that govern autonomous $C^3$ integration to improve the efficient operation of individual robotic elements for the holistic goals of system-wide safety, resilience and reliability. Figure 2 also shows how the continued design of our SDA will result in the identification of new priorities via common behaviors and will drive the evolution into Tier 2 '*Adapt and Thrive*'.

#### A. Symbiotic Interactions and $C^3$ Governance

Symbiotic interactions concern informal and formal relationships that operate under $C^3$ governance. In human-robot systems, it is the integration of human and RAS/I service delivery that creates interconnected strategies in trusted autonomy, augmented learning processes, problem solving and decision-making. These technologies also typically only include one element of symbiosis, as represented within Figure 3.

Symbiotic interactions include the interrelationships between the symbiont and host; we define a symbiont as a system element which requires a type of interaction between another system element to operate. The host is defined as an element with a resource required by the symbiont [91]. The most basic symbiotic interactions are displayed in Table III.

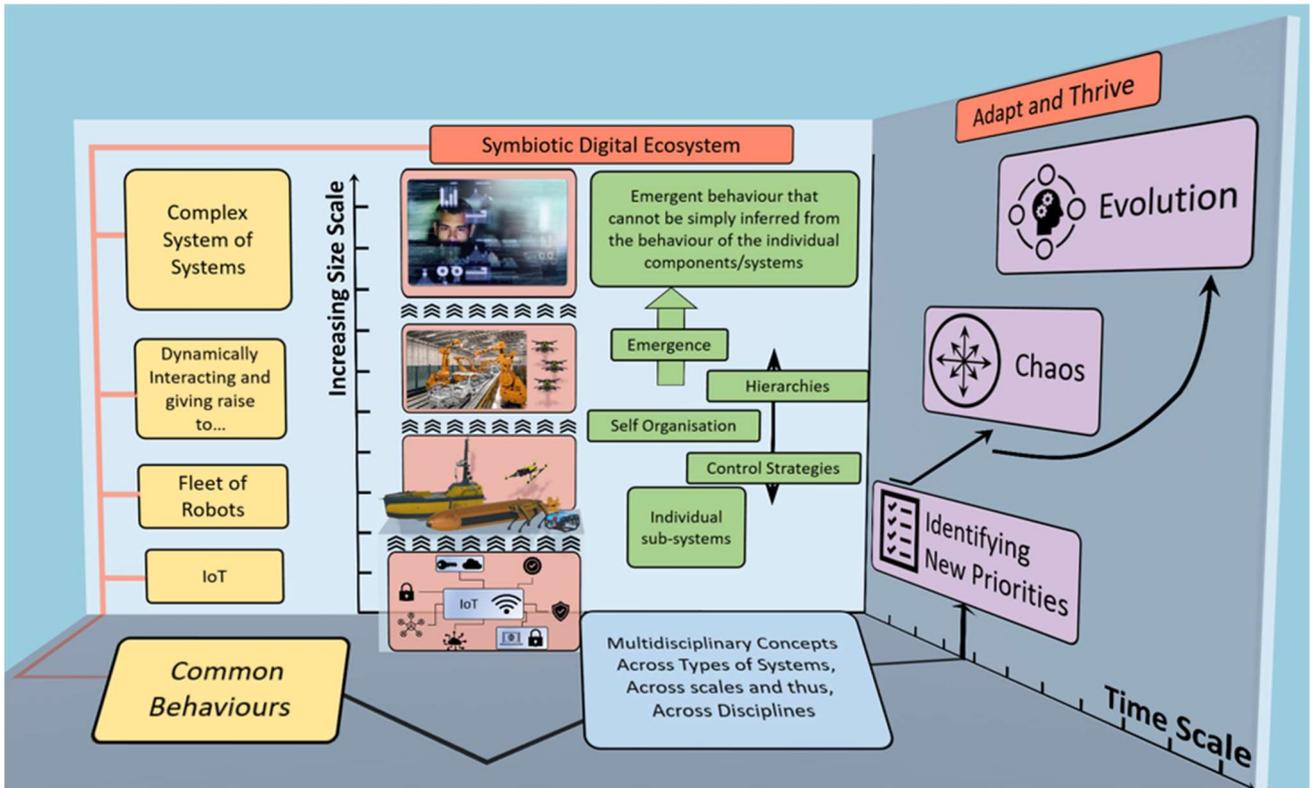

**Figure 2** Diagram of the Symbiotic Digital Ecosystem and the hierarchal steps required to achieve $C^3$ integration.

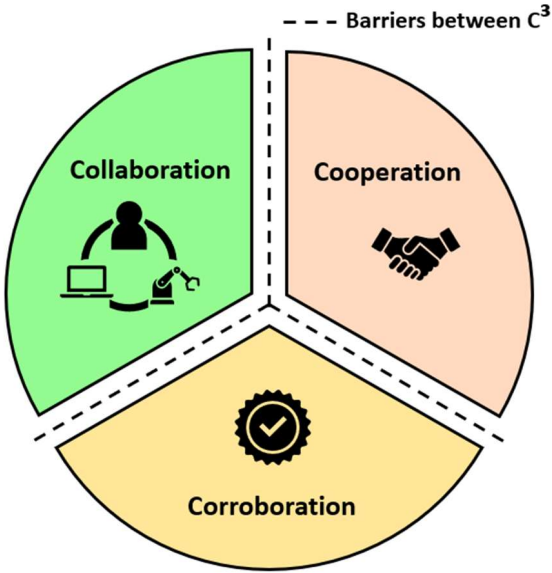

**Figure 3** Barriers to achieving symbiosis across systems.

TABLE III
SYMBIOSIS TYPOLOGY AND FITNESS OUTCOME [125]

| Type of Interaction | Fitness Outcome | |
|---|---|---|
| | Symbiont | Host |
| Mutualism | Positive | Positive |
| Commensalism | Positive | Neutral |
| Parasitic | Positive | Negative |

*Mutualism* is when both the symbiont and host benefit, creating a positive outcome. Examples of this often include the interaction between a human and robot; where the human benefits due to the automated robot completing tasks, and the host benefits as the human can advise the robotic platform of operations.

*Commensalism* is defined by the symbiont receiving a positive result with the host unaffected. An example would be an AI bot improving human efficiency but receiving no benefits in return.

*Parasitism* is represented by interactions between technologies, especially when there is a mix of legacy and new systems, which compete for the same resource, such as power. This may result in the symbiont benefitting at the expense of the host. An example is where a robotic platform (symbiont) connects to a host to recharge its battery to complete a mission and leaves the host with a reduced capacity to complete its own mission.

Teleoperation is representative of a mutualistic relationship, where the visualization assists the human to enhance the operation of the robot, benefitting both elements of the relationship. Whether by visual line of sight and/or a computer display, real-time information is paramount. Today, LiDAR sensors generate real-time maps of the working environment or perform real-time path planning, where the use of mixed reality devices are used to provide new ways of visualizing the local robot, representing a mutualistic relationship [92]–[94].

In the case of speech recognition and DT interaction, commensalism is formed between the operator and the robot. The computational burden of speech processing is on cloud computing infrastructure, so does not affect the performance of the robot, while enhancing the ability of the operator.

Similarly, gesture input is also a cooperative technology, which enables the use of portable mixed reality devices, such as the Microsoft HoloLens, to operate a gesture-based interface. This represents commensalistic corroboration as the computational expense is found on the HoloLens side, which has a dedicated processor for gesture processing. Gestures enhance operator ability, whilst not being detrimental to the performance of the robot. In the offshore renewables context, the human operator can interact with the simulation by viewing the results from the DT of the offshore wind farm without having an impact on the result or operation of the wind turbines. Corroboration occurs due to the comparison of results in the twin model against the real-world platform. Combining this approach with reliable, multimodal input systems should ensure trust in the system is maintained. The integration of corroboration and collaboration is achieved in a DT as the human can, in simple scenarios, stop the robotic platform at any point during a mission. This can be achieved via actionable information from the twin. Utilizing mutualism and commensalism between operator (symbiont) and robotic platform (host) ensures there is no degradation in performance of the mission on either side.

From our literature review, we identify a knowledge gap in the future integration of robotics under a symbiotic system envelope. Current systems are limited to achieving a single element of either cooperation, collaboration or corroboration and do not possess the capability to advance all three of these elements in a single architecture. Consequently, to advance symbiotic interactions, seamless integration and interaction between $C^3$ will result in $C^3$ governance between systems, personnel, infrastructure and environment.

Other examples of symbiotic relationships can be grouped into the following categories, as displayed in Table IV: A symbiotic relationship, which includes a human collaborator, consists of a partnership between a human and robotic platform. This could include safety features to ensure robotic platforms maintain distance from humans or can work in a shared workspace. Multi-platform partnerships can be achieved between robotic platforms to create symbiosis, often achieved in robotic swarms. Infrastructural sensors are often paired alongside DTs of buildings, often including the IoT, and other smart sensors, for a holistic overview of a building, which could include climates, access areas and autonomous systems. DTs are also paired alongside asset integrity inspection devices, where sensors are utilized for structural health monitoring, and where the diagnosed faults are displayed in the digital synthetic environment to be viewed by a remote operator rather than onsite.

TABLE IV
EXAMPLES OF SYMBIOTIC RELATIONSHIPS.

| Symbiotic Relationship | Reference Works |
|---|---|
| Human Collaborator | [126]–[130] |
| Multi-Platform | [129], [131]–[133] |
| Infrastructural Sensors | [134]–[136] |
| Asset Integrity Inspection | [8], [95], [137] |
| System of Systems | [138]–[140] |

To summarize this section, we have identified barriers that prevent full $C^3$ governance in current systems and prevent RAS from achieving our capability criteria stated in Section II. We have summarized the state-of-the-art in technological bridging of mutualism, commensalism and parasitism, via gesture inputs, augmented reality and speech recognition technologies. Section IIIB will describe how our novel SSOSA provides the potential to define complex systems within the systems engineering community combining relationships as in Table IV.

### B. Design of a Symbiotic Digital Architecture for Safety Compliance and Resilience in Autonomous Missions

Safety is a specific challenge in robotics. Although there are many standards deemed relevant by regulators for robotic systems, none of the general safety standards, such as those for industrial robots, collaborative robots, aerospace, and ethical aspects, address autonomy, wherein systems make crucial and safety critical decisions.

In Section IIC and Section IIIA, we have discussed the limitations of current symbiotic systems and system of systems approaches in terms of ontologies and symbiosis, respectively. To address these limitations, our SSOSA includes a range of current symbiotic and novel relationships, displayed as slices in Figure 4, and which further advance the state-of-the-art in symbiosis. The previous development of symbiotic relationships has been instigated from the advancement of DTs, where humans can $C^3$ interactively with the DT, represented as slice A. Most symbiotic systems are recognized as multi-platform, where symbiosis is achieved through collaboration or cooperation of multiple robotic platforms, as represented in slice B.

Corroboration is often achieved through infrastructural sensors, as represented in slice C, where the sensors are used for localization to verify the position of a robot relative to its surroundings. Slice D represents asset integrity inspection, where data sharing and cooperation with DTs present asset faults to the end user. Asset integrity inspection is a less developed symbiotic relationship and is advanced in this paper. Slice E represents our novel symbiosis, both across and onboard the systems of robotic platforms, utilizing bidirectional communications to assess mission status and, crucially, provide self-certification for the wider synthetic environment.

With this view, previous 'symbiotic systems' can be defined as symbiotic relationships due to a partnership between two subcomponents, such as a DT and another subcomponent, as in slices A-C. As identified in Section IIIA, these typically only focus on one element of collaboration, cooperation <u>or</u> corroboration. Hence, our SSOSA encompasses *all* symbiotic relationships, achieving $C^3$ governance to a single DT, as highlighted by the blue shading in Figure 4. These capabilities will be demonstrated using a single robotic platform to highlight the safety and trust created via the self-certification in our system of systems methodology but is also transferrable to other robotic platforms and environmental sensors under the same framework. This transferability will allow us to demonstrate our challenge-based engineering concept and determine its effectiveness in our autonomous mission evaluation.

In the coming sections we present a SSOSA to resilience in autonomous missions as described in Figure 5. We achieve symbiosis across systems within a robotic platform and with the DT by utilization of bidirectional communication throughout our framework for real-time data representation. We have defined a symbiotic system as the lifecycle learning and co-evolution with knowledge sharing for mutual gain. We also define "system of systems" as a set of systems, or system elements, which interact to provide a unique capability that none of the constituent systems can accomplish on their own. This approach is aimed at improvements in operational situational awareness via bidirectional knowledge exchange from a DT, which will optimize performance and encourage life cycle development. This can be completed by aggregating information from across the infrastructure, environment, robot and human-in-the-loop.

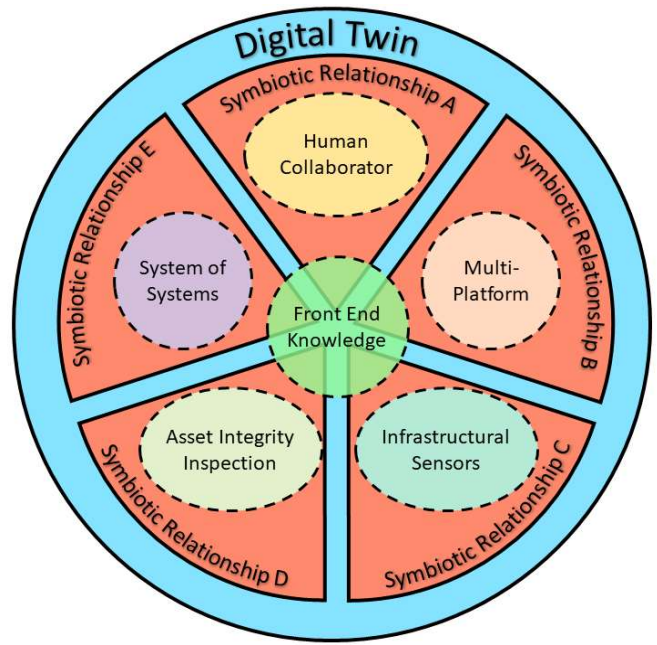

**Figure 4** Symbiotic digital interactions which highlight the integration of symbiotic relationships into our SSOSA definition.

The challenges which symbiotic systems face relate to the creation of a collaboration interface to facilitate trust for the human-in-the-loop. The provision of an improved autonomous system overview to classify mission status, system certification and data sharing without overwhelming the human-in-the-loop represents an additional challenge. The DT component of our SSOSA was designed to act as the command and control of a mission, however, is yet to be implemented to trigger the mission start. The created solution facilitates functionality, human trust, increased autonomy, operational resilience and compliance-certification. Our SSOSA captures many benefits due to the scalable, adaptable and platform agnostic SDA, which features bidirectional communications for increased transparency in operational decision support. This framework can be applied to any COTS platform, where the SDA and system integration process would be adjusted accordingly to suit the COTS platform.

Our Tier 1 SSOSA currently encompasses mutualism and commensalism, which enables the symbiotic architecture to evaluate the implications of a scenario of variables from the environment, infrastructure, human interaction and robot reliability via our run-time reliability ontology. The system also incorporates data sharing from several different sensors deployed in the field, where information is fed to the DT user interface to allow collaboration with a remote human observer. This mitigates known and unknown threats to the resilience and safety case of the autonomous mission. The survivability of the robotic platform is validated due to the completion of mission objectives whilst ensuring continuous safety governance.

Due to the rapid evolution of some failure modes, we designed a recovery strategy that directs the robotic platform to proceed to a safe and accessible recovery zone in the event of an impending failure or warning fault.

For safety compliance we implement mutualism, as the human and robot have the ability to communicate and interact through bidirectional knowledge transfer. This is achieved via human access to the DT to assess the mission status whilst the robot simultaneously prompts the human-in-the-loop with fault diagnoses in real-time. The ontology provides the robotic platform with the capability to autonomously perform mission status decisions, however, the human may stop the mission at any point during run-time.

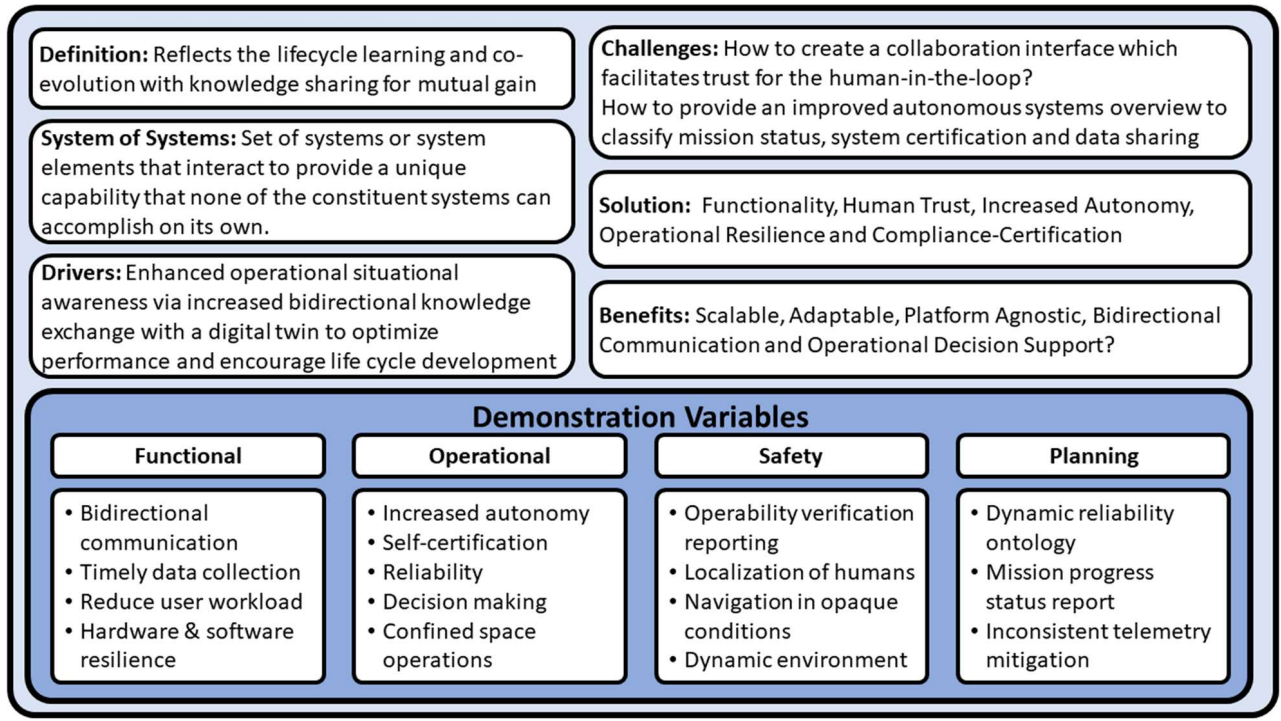

**Figure 5** SSOSA description highlighting the definitions, drivers, solution, challenges, benefits and demonstration variables.

A SSOSA must be resilient during a mission, therefore, our system architecture has been rigorously tested during our autonomous mission evaluation, which consisted of functional, operational, safety and planning variables. These variables are listed in Figure 5, where bidirectional communication and timely data collection for a fully synchronized system is accurately reflected in the DT. The operational variables include increased autonomy and resilience due to the self-certification of systems onboard a robotic platform, where the mission selected was a confined space operation. Safety is a variable which requires operational verification reporting, supports localization of humans and navigation in opaque environments. Lastly, planning ensures that the symbiotic system is autonomous through the design and application of our run-time reliability ontology, which acts as the decision-making hierarchy for autonomous systems and mission progress status reporting to the DT.

The system integration process of our symbiotic system is represented within Figure 6, which illustrates the features of the subcomponents of the system, presents the autonomous mission evaluation and highlights the resilience and symbiosis across the autonomy within subcomponents due to $C^3$ of data. The color-coding implemented in this diagram as displayed in the legend, will be used throughout this paper to provide a common differentiator between internal and external subcomponents of the robotic platform. The layers display the links between all subsystems and highlight the mission variables being addressed. The human-in-the-loop represents the human interaction layer, where the operator can interact with mission components within the DT. The DT represents the user interface layer, which contains the tools and functions for the human-in the-loop to receive an overview of any autonomous systems. The DT is connected to the FMCW sensing data, which is utilized in the confined space autonomous inspection mission. The decision-making Planning Domain Definition Language (PDDL) layer of the run-time reliability ontology is linked to the key software systems of the robotic platform. The decision-making is linked to the Simultaneous Location and Mapping (SLAM) stack, motion planning and ontology. The ontology processes diagnostic data from the internal sensors of the robotic platform. The SLAM stack receives data from the LiDAR sensors and cameras. The motion planning layer calculates the commands to be sent to the mobile base and manipulators. Our system integration process strengthens resilience as each subcomponent, when operating individually, would be unable to resolve the solution required. However, with $C^3$ governance acting across all systems, and our system of system definition, mutualism is achieved across the symbiotic relationships. To support resilient autonomous missions, we focus on the integration of the top-down requirements, as well as the ground-up capability challenges. For information to be actionable within a time critical context, it must be mapped into a design for resilient systems through our SDA, as in Figure 7, which further highlights the functional, operational, safety compliance and planning requirements, and enables resilient symbiosis between a range of systems that are *intra* to robotic systems and *inter* between other robotic platforms.

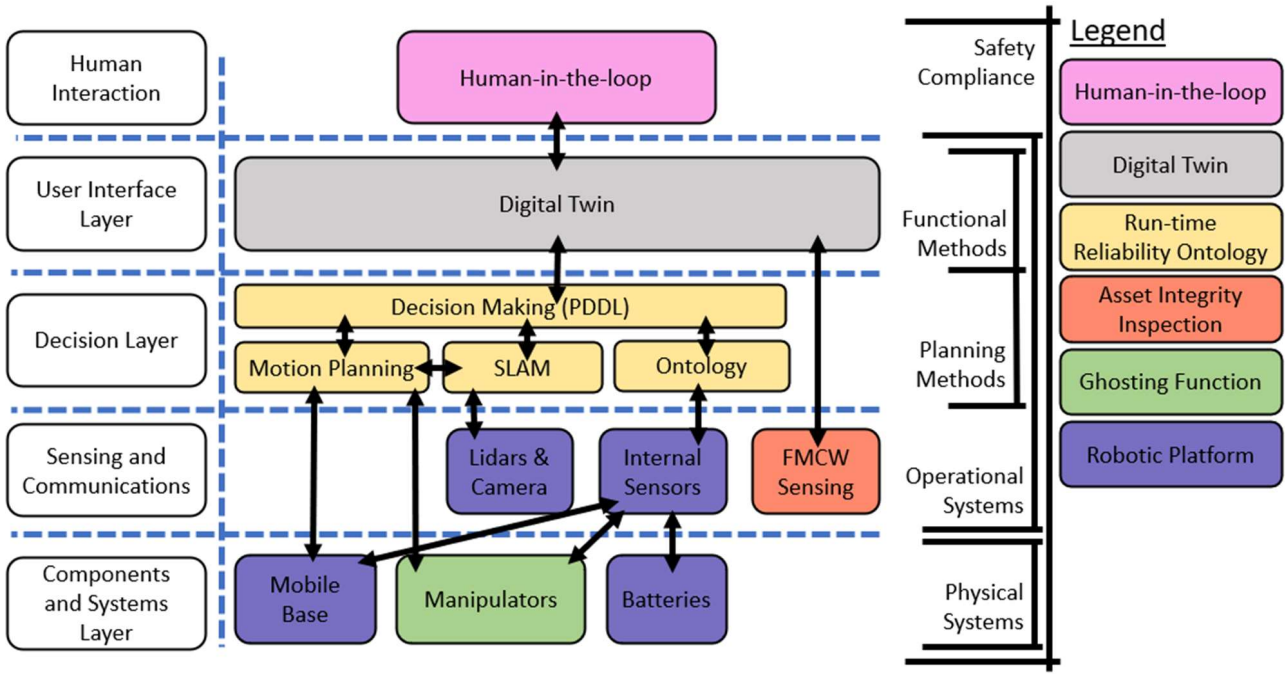

**Figure 6** System integration process of the SSOSA for the robotic platform utilized in the autonomous mission evaluation.

The SDA incorporates the systems engineering which allows the implementation of up to 1000 different sensors and actuators within our architecture [77].

Our SDA commences by supporting the remote human observer at the human-in-the-loop icon and allows the operator to attain actionable information via a bidirectional and interactive GUI within the DT synthetic environment. This information can be acquired via mixed reality devices, resulting in an enhanced hyper-enabled situation report, or via a standard computer. Information about an asset is represented as a digital model of the physical asset in real-time with information, such as defective components, displayed visually via color coding on the digitalized asset.

A meta-function of the DT includes a ghosting function, which increases safety by reducing the risks associated with the manipulation operations. As such, a remote operator can access a visualization of the trajectories of the arms, which can be simulated before being deployed on the real asset. This reinforces trust between a remote operator and the deployed systems, as the operator may visualize intended manipulator actions before committing the movements to the robot, providing increased assurance that the manipulations will be successful and safe. The DT can be used to access machine learning and Multiphysics modelling programs.

For the inspection aspect of our mission, we integrate FMCW radar to provide asset integrity inspection via surface and subsurface analysis. This novel application of FMCW includes detection of corrosion, on metal structures and integrity inspection of composite structure wind turbine blades [8], [18], [95].

Our run-time reliability ontology was developed and implemented to support adaptive mission planning, enabling front-end resilience, run-time diagnosis, prognosis and decision-making. This aids remote human operator understanding of the state of health and remaining useful life of critical sub systems before and during a mission. Our ontology was designed to feed front-end data analysis and edge analytics into these back-end models within the DT. To support connectivity and responsiveness across systems, we synchronize the bidirectional communication modules and data streams for these front-end systems within the DT environment. This includes data from actuators and motors, which are translated into actionable information within the DT, when passed through the ontology.

For each critical part of the system in the AI-driven ontology, a diagnosis automaton is constructed, such as motor, battery, motor driver, wheel, single component or an integrated device, whether sensed or non-sensed. A segment of a system might have its own distinctive states to ensure that rules are in place to govern $C^3$ and the safety of the robotic platform [70].

$$States = \{sensed, possible, normal\} \quad (1)$$

$$Sensed\ states = \{low\ current, high\ temperature, ...\} \quad (2)$$

$$Possible\ states = \{broken, aging, degrading, abnormal\ behavior, ...\} \quad (3)$$

$$Normal\ states = \{on, off, ready, working, ...\} \quad (4)$$

Events that change the states of the components can be internal, temporal, spatial, or external (expected events with different degrees of possibilities). Events on the transition are:

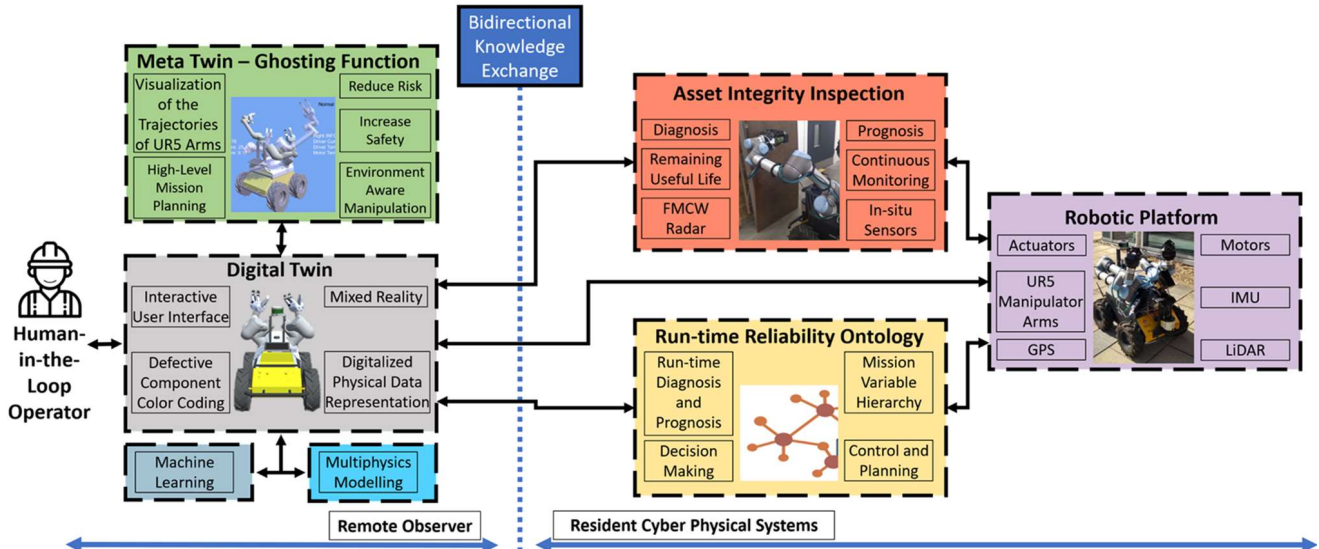

Figure 7 Design for symbiotic digital architecture highlighting subcomponents from human-in-the-loop operator to the robotic platform via bidirectional knowledge exchange.

$$Events = \{internal, time\text{-}driven, space\text{-}driven, external\} \quad (5)$$

A *hierarchical* relationship is used to express all models within the ontology. There are two models namely, *"is-type-of"* and *"is-linked-to"* or *"is-connected-to"*. For example, *"x is-connected-to y"* [70].

$$Binary\ relationships = \{causality, implication, prevention, hierarchical, composition, aggregation, optional\} \quad (6)$$

The logic behind the binary relationship is expressed in the ontology to enable $C^3$ across the subcomponents in the SSOSA. A detailed formalism of the logic can be found in Zaki *et al.* [70].

Three binary relations: '*causality*', '*implication*', and '*prevention*' are combined in modality to show the degree of certainty in the relationship [70]. For example, *x might-cause y, x must-cause y*. Modal verbs combined with those relations include:
- *must* (absolutely certain)
- *would* (really certain)
- *should* (very likely)
- *might/may* (possibly)
- *could* (less possible)

Each part has its own properties which can affect the intra-inter relationships between the parts of the system such as: '*dependency*', '*reusability*', '*validity*', and '*availability*'. For examples, x (is) *stand-alone*, x (is) *reusable*, x (is) *valid*, and x (is) *available* [70]. To summarize the steps:
1. A diagnosis automaton is constructed for each critical part of the system.
2. Describe the transitional relationship between the states.
3. Describe the binary relationship between the states in different components or,
4. Build the hierarchical model of the specific system.
5. Build the generic model of the components.

Key metrics for the ontology presented within this research are the complexity and scalability of the system. For the ontological complexity, our applied system shows that the space requirements are approximately 25 times the size of the raw data, with a linear relationship observed between these two variables. For the reasoning time, the relationship with ontology size begins as an exponential before establishing a linear relationship beyond a threshold value of an ontology size of ~3MB, where the reasoning time is approximately 15ms. A detailed description of these relationships is provided in Zaki *et al.* [70].

The SSOSA presented in this section leads to advances in safety compliance and resilient autonomous missions. An enhancement of operational situational awareness can be achieved via increased bidirectional knowledge exchange with a DT to optimize performance. The subsequent design of our systems-engineered SDA will lead to scalable and transferable platforms sector-wide, while meeting the capability criteria. The further design and development of our SSOSA will lead to improved capabilities in cyber physical systems and the servitization of fleets of RAS and compliments the need for future applications of trusted BVLOS systems.

## IV. IMPLMENTATION

### A. The Demonstration Description

An onshore training facility configured to resemble an offshore substation platform was used to evaluate the integrated symbiotic robotic platform and the accompanying synthetic environment [70]. Key features of a typical offshore confined space, such as offshore generator or high-capacity transformer room, include complex arrays of piping and cabling and very large infrastructural elements. Although protected from weather, the ambient conditions and the electromagnetic environment posed wireless telemetry challenges. As part of our mitigation strategy to ensure reliable communications during BVLOS operations in confined spaces, a wireless base station, paired with high gain wireless transceivers onboard the robotic platform, was employed.

The test area consisted of many obstacles over and around the transit route. The identified constricted areas had minimal clearance on each side of the robotic platform, resulting in regions of significantly increased collision risk. Path parameters were tuned to allow for high performance during the confined space navigation stage, whilst still maintaining collision avoidance. This offshore platform analogue represented a highly challenging environment for sensing and high-fidelity SLAM functions.

### B. Mission Description

The mission was partitioned into eight distinct stages which contain key mission waypoints:
A. Pre-mission planning
B. Mission start at base point
C. Transit to asset integrity scan 1
D. Perform asset integrity scan 1
E. Transit to asset integrity scan 2
F. Perform asset integrity scan 2
G. Return transit to base point
H. Mission end

Stages A-H represent the full execution of the asset integrity inspection mission. In addition to the core mission waypoints and objectives, three major system issues were included to simulate symbiotic collaboration dynamics, which are: the necessary symbiotic reassessment of the mission by the robotic platform, intra-inter system self-certification and adherence to safe operational protocol. This validates our '*adapt and survive*' paradigm, where the dynamic conditions imposed on the mobile robotic platform create the need for symbiotic AI-assisted decision-making in commensalistic collaboration with the system reliability ontology. In alignment with our three capability criteria from Section II, this ensures the robotic asset can:
- Identify threats or barriers to the success and safety of the mission via integrated sensing.
- Provide run-time cooperation with a DT system to relay acquired asset integrity data and to inform parallel robotic elements and human-in-the-loop operators in run-time via bidirectional knowledge exchange.
- Corroborated decision-making and trusted autonomy through both AI and/or the human-in-the-loop operator via wireless, low-latency communication.

A key aspect of this work is to demonstrate resilience while operating autonomously and entirely within the envelope of safety compliance. This is achieved by the $C^3$ in the SSOSA methodology to provide improved, real-time human-in-the-loop awareness and the symbiotic $C^3$ governance between the systems that allow the robotic platform to operate autonomously and safely.

While the mission envelope is defined as a confined space asset integrity inspection, we also assess the reliability and resilience of the robotic system by inducing randomized faults during the mission. Our reliability ontology facilitates symbiotic evaluation of the robotic platform to self-certify its systems and terminate the mission, if necessary. Figure 8 illustrates the mission plan on a mission area map. Figure 9 shows the mission duration as a function of symbiotic interaction for the asset integrity inspection mission, where no faults occur. Figure 9 (inset) shows the increased level of symbiotic interaction due to simulated faults induced within the robotic system and the altered mission path, demonstrating the SDA response and recovery strategy. An itemized description of the demonstration is also provided in Appendix I. The following subsections discuss the mission description and the different outcomes at each waypoint. The applied methodology during the mission is presented in Figure 13 as a flow chart of the operations, symbiotic decisions and interactions between systems.

*A. Pre-Mission Planning*

Pre-mission planning is critical to the success of any confined space mission. A reconnaissance mission is performed to map the area prior to O&M work to establish the working environment. For this evaluation, an operator manually navigates the robotic platform around the environment and infrastructure to create a map before adding the waypoints (Figure 13A), representing cooperation between robot and human. To ensure that raised surfaces, such as pipework and low obstacles were detected, a 3D LiDAR mounted above the body of the platform, in combination with a 2D LiDAR, mounted low on the platform (Figure 10), provided the SLAM data for the DT. The resulting reconnaissance map is displayed as the floor plan schematic used in Figure 8. For the autonomous inspection evaluation, the FMCW radar was fitted as a payload on a pan tilt unit.

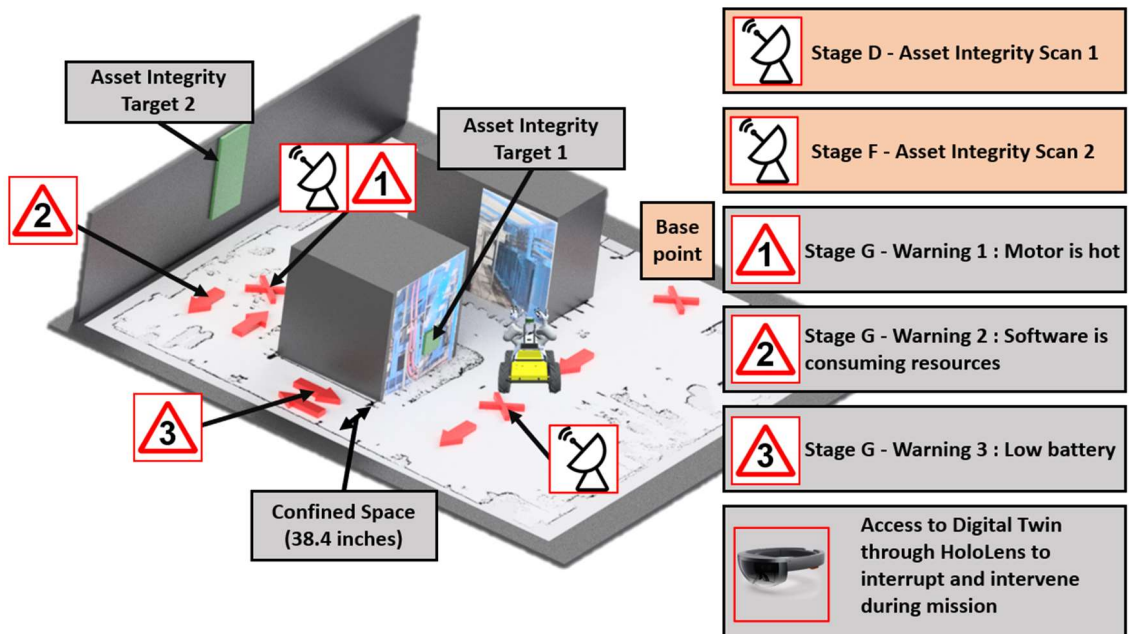

**Figure 8** Mission plan within an 3D model of the industrial facility highlighting key stages of the mission events and route taken by the robotic platform.

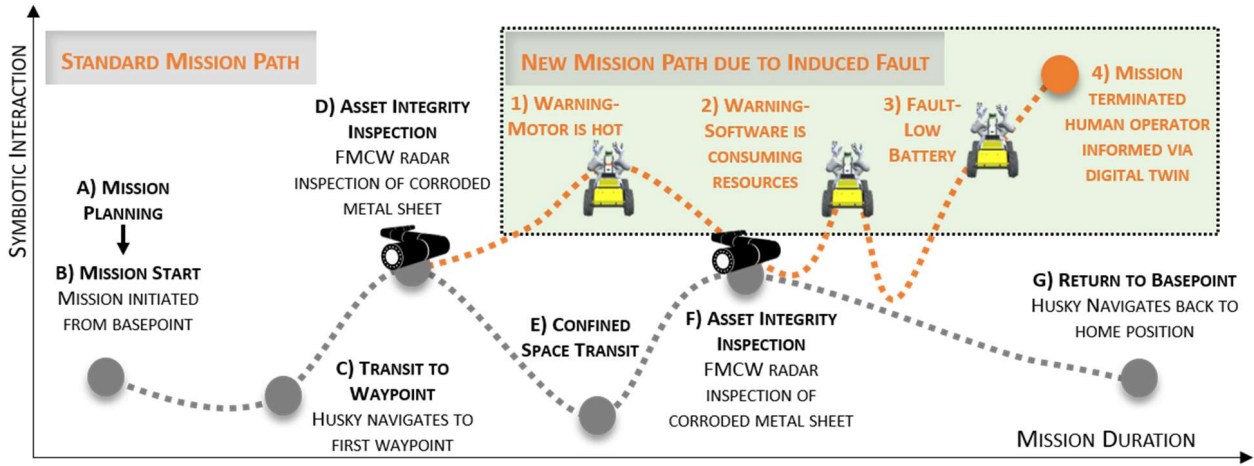

**Figure 9** (Main) Stages A-G represent key events on the standard asset integrity inspection mission path. (Inset) 1-4 represent systems warnings and faults resulting in autonomous symbiotic interaction for mission continuity or the safe recovery of the robotic system.

A robot equipped with LiDAR sensors, compared to onboard stereo cameras, has the following advantages; computationally inexpensive processing relative to stereo image processing demands, generally longer range, improved accuracy with less noise and functions independently of environmental lighting conditions. The interaction created from using these types of systems enable robots to actively complete objectives, where camera-based systems simply observe.

For the demonstration, the ROS navigation and planning stack was used. Decision-making, based on PDDL, ensures relational sequential system actions are achieved where the robotic platform cooperates with the assigned tasks by the operator [96]. One form of action is a waypoint goal. These waypoint positions are passed to the navigation stack by the planner, where the SLAM data is used for navigation. Movement between these waypoints is handled by a ROS *move_base* navigation stack during run-time. The DT provides interaction for an experienced operator/planner to create waypoints. From the reconnaisance mission, an accurate and effective map of the area is created, ensuring that the selected robotic platform is capable of completing the required mission.

### B. Mission Start at Base Point

The robotic platform remains idle at an approved base point until triggered by the operator. This requires reliable wireless connectivity between the DT and robotic platform. From the moment the mission is triggered, the system actively self-certifies its systems (intra-system corroboration) via watchdog nodes, which are subscribed to fault data from the ontology. This ensures system deployability is visible to the human operator via the DT. The DT serves as a real-time collaboration hub, where the underlying methodology is represented in Figure 13B. The autonomous navigation and mapping systems are initiated to ensure the mobile platform computes the most efficient route to complete the mission.

### C. Transit to Asset Integrity Scan 1

SLAM, in conjunction with the low-level path planner, is used to reach the first waypoint, where a global costmap is used alongside a live updated local costmap during the mission. The global costmap represents the map generated from the pre-mission planning stage. The local costmap represents data collected live from the LiDAR systems as seen in Figure 11. In the grid, cells are marked as 'clear' or 'occupied' using points detected by the onboard LiDAR systems. The integration of both costmaps enables corroborative navigation to reduce the risks associated in autonomous navigation. The PDDL planner outputs a waypoint goal action containing x, y and θ positions, as input into ROS *move_base* for autonomous navigation (Figure 13C).

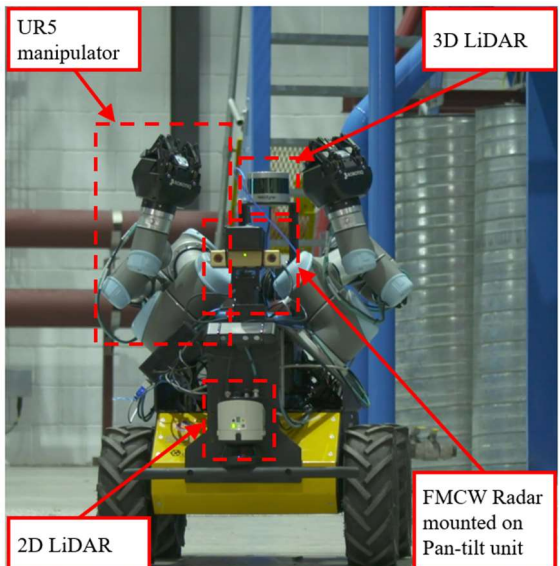

**Figure 10** Dual UR5 Husky A200 with annotations for the onboard payloads.

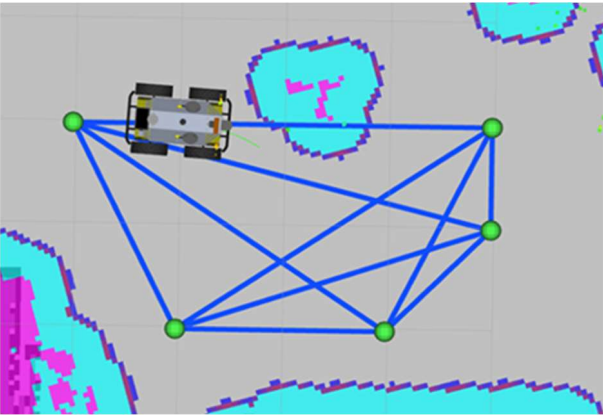

Figure 11 Example of a local costmap highlighting waypoints which are positioned by the human operator.

### D. Perform Asset Integrity Scan 1

The first inspection is completed at this waypoint based on the scan action determined by the generated plan. The FMCW radar sensor was used for non-destructive analysis for corrosion. Operating in the K-band, the sensor acquired return data for 30 seconds, with each chirp lasting 300ms over a frequency sweep of 24-25.5 GHz. The challenges here pertain to maneuvering the robotic platform safely, without colliding with the infrastructure, and that the robotic platform is a safe distance from any infrastructure. This mission objective is presented in Figure 13D and displays $C^3$ governance, exemplifying the corroboration of the result from the inspection, the cooperation in ensuring that the FMCW radar sensor is orientated and cooperation through the adaptive navigation to reach the waypoint. Achieving mutualism for both the robot and remote operator.

### E. Transit to Asset Integrity Scan 2

The transit to the asset integrity scan waypoint, which is the location of the most constricted access in this confined space mission. This area is classified as a hazardous zone, as the route features a narrow entry to the asset integrity scan area. For this demonstration, the base motion path planner was configured to navigate through confined spaces whilst still avoiding collisions, enabling the robotic platform to cooperate with its environment to ensure safe entry.

### F. Perform Asset Integrity Scan 2

Upon arrival at the second waypoint, the robotic platform performs the asset integrity inspection autonomously. The functional, operational, planning and safety challenges are very similar to the first asset integrity inspection ensuring safe maneuvering of the manipulator arms and of the robotic platform (Figure 13F).

### G. Return transit to Base Point

Three induced faults were simulated on the robotic platform via additional code activated within the core. The fault severity levels are classified as in warnings 1-3 (Figure 8 and Figure 9). One of the key impacts of our research relates to the fault detection and warning thresholds set by our novel ontology, which qualitatively improves the resilience in the systems as this information is transferred from the SDA to the DT, enhancing an operational overview. These improvements include autonomous detection of onboard faults via the ontology for self-certification and shared knowledge exchange to a digital twin for a remote operator.

To identify faults and support run-time diagnosis of the autonomous systems, a formal representation was utilized. The ontology formalism is comprised of different sets of semantic relationships (mutualistic) and diagnosis automata to model the system. The relationships between the components are made at the top-level between the components, or at bottom-level between the different states of the components. A diagnosis automaton is constructed for each critical part of the system, i.e., stand alone or integrated devices, whether it is sensed or non-sensed [70]. Different states can be attributed to specific system elements. The model is initially based on a hierarchical relationship, where classes and subclasses are displayed showing the required detail for an accurate ontology model. The object properties include the parameters of each variable, which must be allocated to ensure faults are detected by the AI-driven real-time reliability ontology. The ontology ensures the cooperation with the subcomponents in the system; DT and robotic platform. The ontology mutualistically assesses the state of health of the robotic platform. If a warning is detected, the ontology relays the results via $C^3$ to the human operator. The bidirectional communications enable cooperation and collaboration via the interactions between the human operator and DT. For example, if a warning is presented to the human operator via the DT and the robot has autonomously continued the mission, the human operator may still terminate the mission.

We recognize that deployed robots will develop malfunctions and faults within their systems. Consequently, our main goal is to detect or discover anomalies or invalidities in the system under stress. The end objective of the run-time reliability ontology is to validate that the behavior of the robot matches the required specifications (corroboration). Four test cases are considered:

- A possible problem in a non-sensed component, for example, a wheel.
- Prediction of low battery voltage.
- Root cause analysis for two components affecting a third.
- Prediction of high temperature in the motor driver.

The three warnings induced in the system, alongside their challenges, are displayed within Appendix I-Stage G, where the implementation of AI via the ontology prioritizes fault thresholds over warning thresholds in all cases to ensure the integrity of the robotic asset. The novel procedure incorporated within our decision-making algorithms is represented within Figure 13G to identify the interactions of the SDA and system integration process between each system.

*Warning 1* denotes increasing motor temperature towards preset warning thresholds, where the motor is still functional. The relationships which represent the detection of the motor temperature increase are shown in Algorithms 1 and 2 (Appendix II). A prompt to the human-in-the-loop is only required at this point in the mission to notify of possible overheating, with consequential effects on functionality, operation, planning and safety. If the warning condition persists, the human-in-the-loop is presented with the option to terminate the mission. In the demonstration, the real-time reliability ontology autonomously notifies the human operator whilst continuing with the mission.

*Warning 2* pertains to computational process management. This is exemplified via management of the limited computing resources, which could result in other data processing and control being delayed in the event of an error (parasitism). Consequently, this results in longer computation time and delays in mission critical software processes. The run-time reliability ontology utilizes the pseudocode in Algorithms 3 and 4 (Appendix II) to detect if the Random Access Memory (RAM) or Central Processing Unit (CPU) is consuming the resources. The human operator is once again prompted with a warning while the mission continues.

*Warning 3* alerts the human operator to a low battery/State of Charge (SoC). This is a critical situation for the robotic platform, as reduced current availability requires replanning of mission capabilities. Under our '*Adapt and Survive*' paradigm, the ontology executes the decision to prevent further degradation to the robotic platform. The management of safety has been considered as the integrity of the platform is compromised, however, still recoverable by a human. Algorithms 5 and 6 (Appendix II) represent the fault level threshold and warning threshold, which allow the ontology to identify when the SoC of the battery is low. Lastly, the human-in-the-loop is well informed via the DT and has an accurate prognosis of the system status.

### H. Mission End

During this mission stage, we demonstrate the benefits of our run-time reliability ontology. Warnings were detected on the route of the mission, where the ontology had the option to terminate the mission autonomously for each consecutive error. Each warning also allowed the human-in-the-loop to terminate the mission if necessary. To ensure adherence to safety governance, the robotic platform assesses its ability to operate effectively after each warning, thus ensuring continued survivability and resilience. Many warnings were collected, therefore, to prevent failure and ensure the integrity of the robotic asset, the outcome from the ontology autonomously prevents the mission from continuing and awaits recovery. The human-in-the-loop was informed in real-time of the mission status via the DT interface. The twin presents the representation of data converted to filtered ontology messages, displaying hardware and system faults to the user via a red color-coded alert system. For this mission evaluation, the robot was autonomously stopped by a watchdog node subscribed to fault data from the ontology due to the low battery fault, as in Algorithm 5 (Appendix II). The fault was presented in the DT and represented the system health status, as in Figure 12. The interface was designed to draw the attention of the human operator to the high priority alerts. The DT also presents lower order information, such as battery status parameters. $C^3$ governance ensures a framework of coordination, adjudication, and integration of all the subcomponents, systems and human-in-the-loop goals with a SSOSA. This identifies another key impact due to the symbiosis between the ontology and the DT, where a remote operator can access diagnostic information and the warnings detected by the ontology in real time for a BVLOS system.

The taxonomy structure in Table V presents an analysis of the mission performance via the symbiotic safety compliance modes regarding the motor temperature of the robotic platform. Each safety compliance mode is identified according to their specific $C^3$ governance elements of system awareness, provision, operation and outcome, corresponding to Mutualism, Commensalism and Parasitism (MCP) relationships. The SDA relies on these relationships to create interactions between, or across, the robotic platform, ontology, DT and human-in-the-loop. System awareness includes the ability of the robotic platform to be aware of its own capabilities. For example, system awareness allows platform self-preservation without affecting the human; although the mission has stopped, the integrity of the robot is maintained due to self-certification. In this autonomous mission evaluation commensalism is high, mutualism is moderate, and parasitism is low; the robot continues its mission with a minor possibility of degradation to the robotic platform state of health.

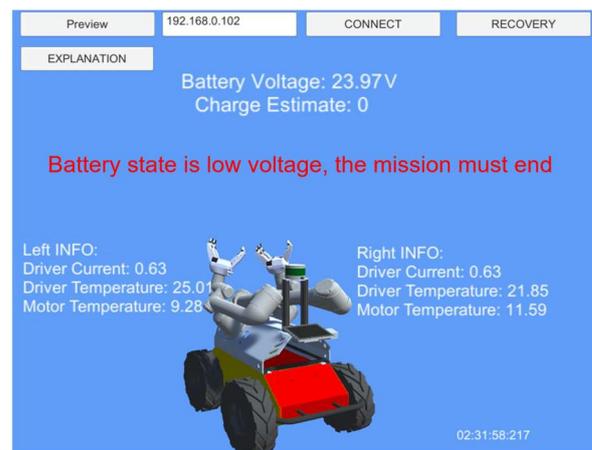

**Figure 12** Low battery error message displayed in the DT alongside color-coded alert system in red on the mobile base indicating the health status of the robotic platform.

TABLE V
TAXONOMY OF SYMBIOTIC SAFETY COMPLIANCE FOR
ROBOTIC PLATFORM MOTOR TEMPERATURE

| $C^3$ Governance | Safety Compliance Modes | | |
|---|---|---|---|
| | Mutualism | Commensalism | Parasitism |
| System Awareness | Moderate | High | Low |
| Human-in-the-loop Provision | High | Moderate | Low |
| Operation | Self-certification (Implication) | Augmentation (Causality) | Instructional (Prevention) |
| Outcome | Positive Anticipation | Indeterminacy | Negative Anticipation |

The reliability ontology ensures that human error is minimized throughout a mission. The fault thresholds for any problems are set such that the robot terminates the mission if any unsafe operating condition is detected. Under human-in-the-loop provision, the ontology continuously conducts state of health assessment, hence, parasitism is low, as shown in Table V. Mutualism occurs when information is used, representing a shared understanding that would not have been possible without each subcomponent augmenting the other. Augmentation occurs at both information and data levels in the SSOSA. Here, the human-in-the-loop is prompted by warnings (information) and a new fault threshold (data) is triggered that instructs the robot to terminate the mission. A balance between commensalism and parasitism can be achieved if an experienced operator alters fault thresholds during the mission planning phase. Commensalism is attained when fault thresholds are further altered by an experienced operator, resulting in mission termination if, and only if, the thresholds have a minimal amount of risk to the mission, as corroborated by the reliability ontology. This can also occur in the scenario where a warning threshold is reached as the human has the option to terminate the mission. In this scenario, knowledge transfer occurs from the ontology to the human operator via the DT. Parasitism, though reduced, can occur if an experienced operator has set the inappropriate fault thresholds for component reliability within the ontology, resulting in a priority over the mission, but to the detriment of the robotic platform state of health.

### C. Scenario Modelling

A multi-level *'Adapt and Survive'* paradigm requires proactive system interrogation and response. Due to the complexity of robotic systems, this can lead to several warnings, faults and failures. We present three scenarios where the robotic platform is to safely return to the base point. The scenarios are designed to verify the resilience of the run-time reliability ontology and therefore the self-certification of the robotic platform. To evaluate the different $C^3$ governance levels of autonomous intervention, a self-certification model was derived from candidate components of the reliability ontology schema. The logic base contains finite state automata for each sensed component and for some of the non-sensed components in the system [70]. This novel approach enables effective runtime diagnostics and prognostics. The results show that the proposed approach and modelling paradigm can capture component interdependencies in a complex robotic system. The resulting artifacts can be processed within 10ms to support front end mitigation, inferring the scalability of the proposed approach.

The three scenarios represented in Appendix I- Stage H are as follows:

*Scenario 1: No warnings or faults detected by reliability ontology - Mission success*
No reliability issues were induced in the system. The ontology operates and verifies the healthy state of the robotic platform. No warnings are prompted to the human-in-the-loop.

*Scenario 2: Warnings detected only - Mission success*
Low-level faults conforming to warning thresholds were induced to the system. The identified problems are within the warning threshold, but have not yet reached the fault threshold, therefore the mission is still achievable. The ontology diagnoses the problem and converts this data into actionable information for the human-in-the-loop. This determines that the robot can continue with the mission but updates the human-in-the-loop so they can determine if the warning has too much risk associated.

*Scenario 3: Many warnings and a major fault detected by the reliability ontology – Autonomous Mission Termination*
Severe faults were induced in the system to verify and validate that the ontology can diagnose problems reliably and accurately during run-time. This mission pertains to resilience, reliability and safety compliance, in keeping with the capability criteria stated in Section II. As several warnings are induced on the robotic platform, the ontology terminates the mission to prevent further deterioration of the robotic platform. This represents an example of parasitism in mutualistic collaboration, facilitating more stable cooperation.

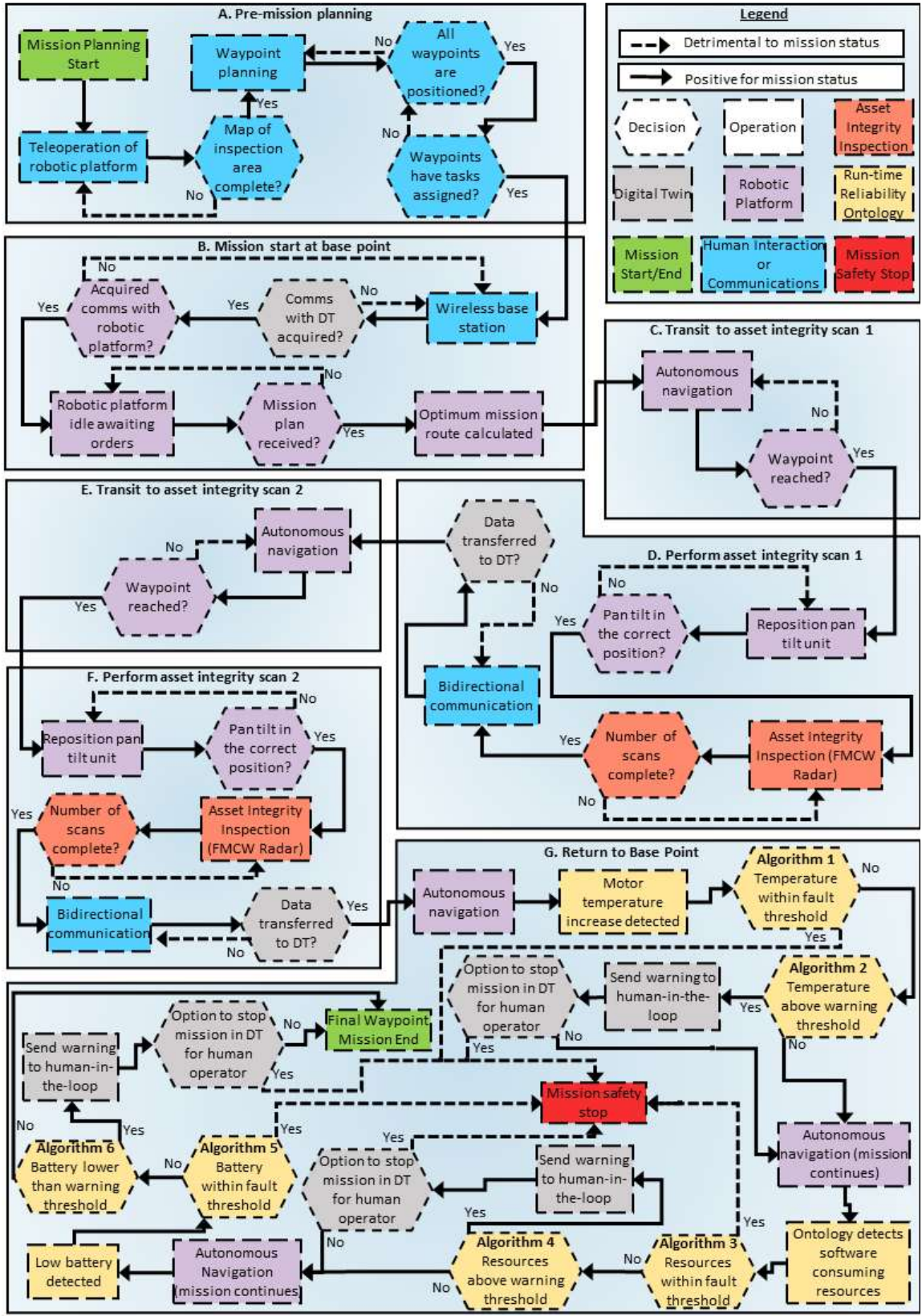

**Figure 13** Methodology of SSOSA during the autonomous mission evaluation highlighting operations, decisions and the system of system interactions. The same color code has been applied to identify subcomponents as in Figure 6.

### D. Digital Twin

A DT is defined as "*digital replications of living as well as non-living entities that enable data to be seamlessly transmitted between the physical and virtual worlds*" [97]. We report a "Stage 4" DT with extended data analytics and simulation capabilities, in particular leveraging edge-processing in real-time to predict future behaviors (Figure 14). A DT designed according to this paradigm ensures positive interdependency across its internal and external functions, allowing integration of real-time sensor data streaming and processing with other operational RAS/I inputs and services. It ensures legitimacy is maintained in and across existing technology ecosystems.

Hastie *et al.* cite three main challenges for human-robot collaboration, which this work addresses: planning in human-robot teams, executing and monitoring a task and adaptivity of the human-robot partnership [89]. Meeting these challenges requires pre-mission planning, situation monitoring with the ability to manually assume control, if necessary, and the ability to re-synchronize with the robot if communications are lost.

The common prevalence of internet connectivity and the increasing number of cloud computing solutions available have enabled the rapid development of cloud robotics [98]. The technology is fundamental to DTs and offers an extremely powerful computing platform without the associated hardware costs. Importantly, it allows ease of integration and communication with edge-devices and robots, including human-robot interfacing. The following subsections describe the functionality of the DT used for this mission evaluation.

#### 1) GHOSTING OF DUAL MANIPULATORS

Our mission evaluation incorporates manipulator capability, which is considered integral to the future of BVLOS autonomous missions. The role of manipulators in the offshore environment is twofold; for the maneuvering or carrying of payloads and the potential manipulation of valves or switches. An emphasis is placed on intuitively informing an operator about the status of the robotic platform and manipulators. Hence, run-time analysis and collaboration features of the manipulators are available through the DT interface, which allows the user to monitor and control the robotic manipulators in real-time. Messages generated by the reliability ontology are displayed and the user can interactively control the manipulators on the robotic platform, mirroring their real-life condition during run-time. The design and development of these functions will become increasingly important as fully autonomous robotic systems are employed in BVLOS roles.

A DT server package integrated into the robotic platform ROS core ensures run-time connectivity between the robot and client machines. The DT interface does not tie the operator to a single ROS-driven machine, and through the SDA an operator can connect via any device, anywhere, and remotely to the robot. The DT GUI provides visualization and interaction, demonstrating the SSOSA and SDA for process control. This is achieved by utilization of run-time prognostics to verify the value of the bidirectional communications. This enhances the interaction via a physics-based simulated operational preview that supports trust and system state of health.

Figure 15 displays the DT ghosting function, which enables the remote planning and control of the manipulators. Trajectories and planned positions of the arms are displayed to the user as a translucent "ghost" model, allowing the operator to preview and analyze the requested operation. Sliders to control the "ghost" arms are provided in the DT

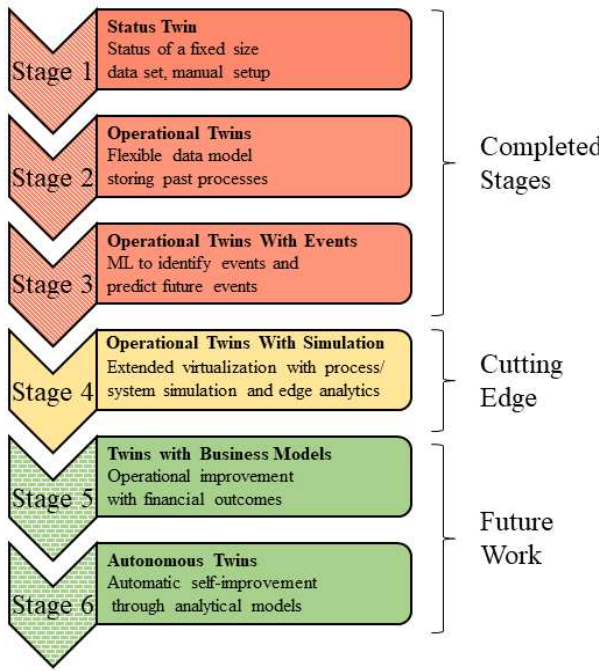

**Figure 14** Stages of a DT indicating Stage 4 as the current model presented within this publication.

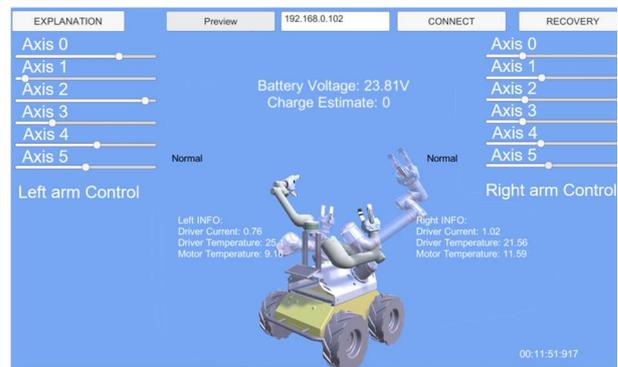

**Figure 15** Meta ghosting function of the DT, highlighting the controls and trajectories of the manipulators.

GUI to simulate each axis of the manipulators. This allows remote operators to verify safe manipulator motions through simulation before committing the execution to the field robotic platform, increasing the level of trust and ensuring the manipulators act as intended.

The DT was also evaluated for run-time fault prognosis, with the arms color coded red as in Figure 16 for the visualization of fault detection within the simulation of movements. For this illustration, we induced a motor fault on the manipulator via the ROS core.

### 2) MIXED AND AUGMENTED REALITY

On-site and remote human-robot collaboration allows rapid assessment of the state of health of a robotic platform via mixed and augmented reality. Figure 17 presents an augmented reality interface where natural language is used to indicate the health status of the robotic platform via the corresponding Quick Response (QR) code. Color coding identifies the health status of components (Figure 18); when viewed through the augmented reality interface by a remote operator, the base of the robotic platform is visually highlighted red for easy remote identification of a fault. The color coding of faults can be tailored depending on the platform and the nature of the fault.

In this section we have identified our position within the multi-stage roadmap required to achieve fully automated self-improvement through analytical models. We have demonstrated that our work represents the cutting edge for implemented augmented reality to meet the human-robot interaction requirements, as stated in our capability criteria in Section II, for edge analytical asset representation, robotic state of health and ghosting of manipulators.

## V. APPLIED MILLIMETER-WAVE SENSOR INTEGRATION FOR OFFSHORE ENVIRONMENTAL SENSING

### A. Asset Integrity

Identified as a key emergent candidate technology for robotic deployment and asset integrity inspection in Section IIE, the FMCW sensor used for this mission evaluation offers fast measurements, with a 300ms chirp duration and 30ms computation time. This facilitates online monitoring via edge analytics, while also providing low sensitivity to environmental conditions and non-destructive evaluation of targets. The driving electronic modules are low power (~900mW at maximum power draw), solid state devices suitable for Atmospheres Explosible (ATEX) compliant areas. Millimeter-wave sensing provides adjustable acquisition rates and is proven to be effective in harsh operating environments, such as high pressure or high temperature areas. Millimeter-wave sensing is also proven to function in an opaque environment, such as fog, mist, dust and smoke [99]–[107].

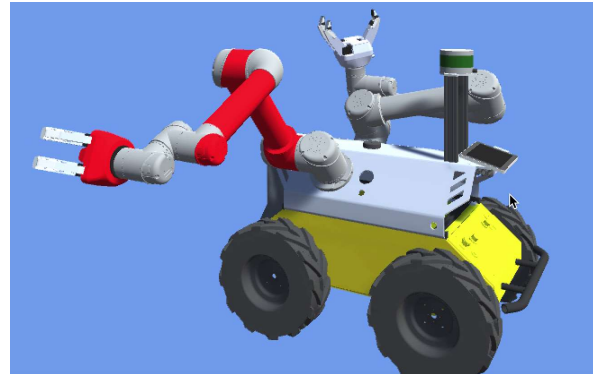

**Figure 16** Meta warning function of the DT, where the arms are color coded highlighting the protective emergency stop in the simulation.

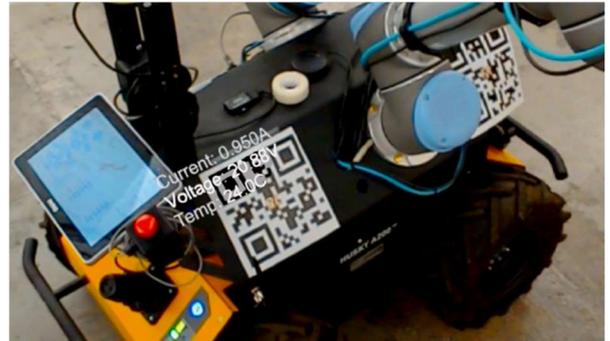

**Figure 17** Mixed reality interface showing natural language of the health status of the robotic platform via the corresponding QR code.

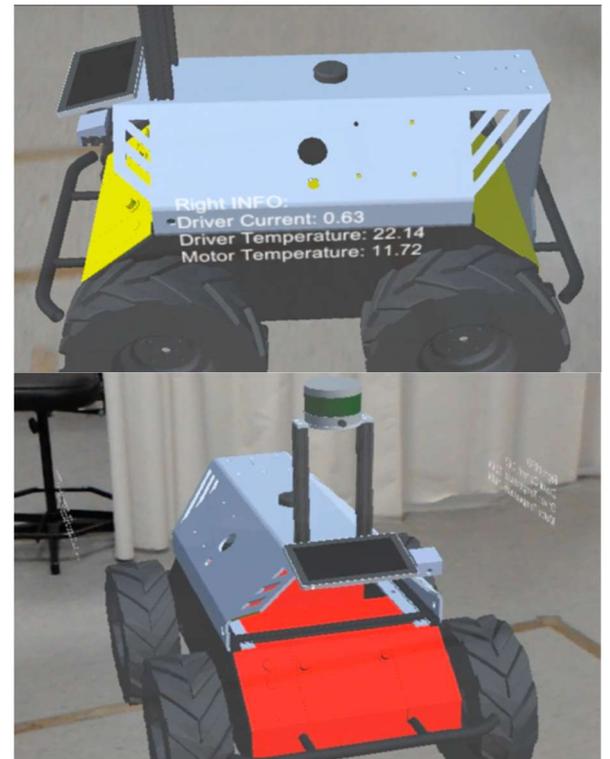

**Figure 18** Augmented reality interface where the top image displays the current health data of the robotic platform and bottom displays a color-coded fault alert indicated as the red base.

Acting as an edge analytical device on the deployed robotic platform, we demonstrate the capacity of the sensor to return critical asset integrity information for application to real-time DT reporting in the offshore asset integrity role. This section details two successful use case applications for the millimeter-wave sensor in the offshore renewables sector: steel infrastructural corrosion assessment and wind turbine blade integrity monitoring. Both use cases utilize the millimeter wave sensor as an inspection device during the mission profile described in Section IV. The implementation of the device further enhances our SSOSA, as the asset integrity data updates the DT providing corroboration of asset health. The manipulator arms allow the robot to perform raster scans with the FMCW unit to assess wider areas for faults. The dual UR5 manipulators mounted on the robotic platform can be tasked with differing objectives, where one manipulator may be used to maneuver the sensor, allowing the other manipulator to perform asset interventions, where necessary (an example of cooperation).

1) STRUCTURAL CORROSION

The detection and quantification of surface corrosion on steel structures is critical to the implementation of O&M schedules in the offshore renewables sector [108]. Figure 19 shows the robotic platform during an asset integrity inspection for corrosion. Figure 20 displays the observed return signal amplitude response for differing targets of metal and concrete at a consistent 10 cm from the sensor tip. Clear order-of-magnitude contrasts between the non-corroded and lightly corroded steel sheet were observed and quantified, in addition to significant contrasts for the polished aluminum and consistent values for differing areas of the concrete test area floor.

The application of corrosion mapping within a DT of an offshore asset improves operator understanding of remote asset health conditions and facilitates bespoke maintenance scheduling. This leads to an enhancement of the relationships across $C^3$ with a significant increase in corroboration of real-time viewing of asset health.

2) WIND TURBINE BLADE ASSET INTEGRITY INSPECTION AND DASHBOARD

In this section the DT framework has been further applied to asset integrity inspection via FMCW radar sensing. This modality provides a user-friendly display of the state of health of subsurface materials that comprise the interior of a wind turbine blade. This allows an operator to not require prior knowledge of FMCW theory and responses and provides an intuitive information display for accessible human interaction (an example of commensalism).

In this section, we utilize a decommissioned wind turbine blade, exhibiting a type 4 delamination defect on the internal structure of the blade, as pictured in Figure 21A and inspection area in Figure 21B [8]. We demonstrate the ability of the millimeter-wave sensor to detect the presence of key defect types and the environmental conditions that accelerate the asset degradation. The subsurface faults are inspected from the exterior of the blade and are represented in our Asset Integrity Dashboard (AID) as depicted in Figure 22, which provides easy access to an operator for information regarding the integrity of their wind turbine blade. The information is easily identifiable due to the color coding, where green represents a healthy section of the blade and red hatching represents a defective area. User interaction, by clicking on the defective area, displays a summary of the fault diagnosis. The operator can view further information, such as the detailed radar response in addition to detailed descriptions of the fault. Figure 23 shows the FMCW sensor response to: an area of undamaged wind turbine blade structure, an area of structure identified to exhibit a type 4 delamination defect and the same area of defect with the addition of 3 milliliters of fresh water [8]. A video demonstration of the AID tool highlights the interactions and results from the FMCW radar [109], [110].

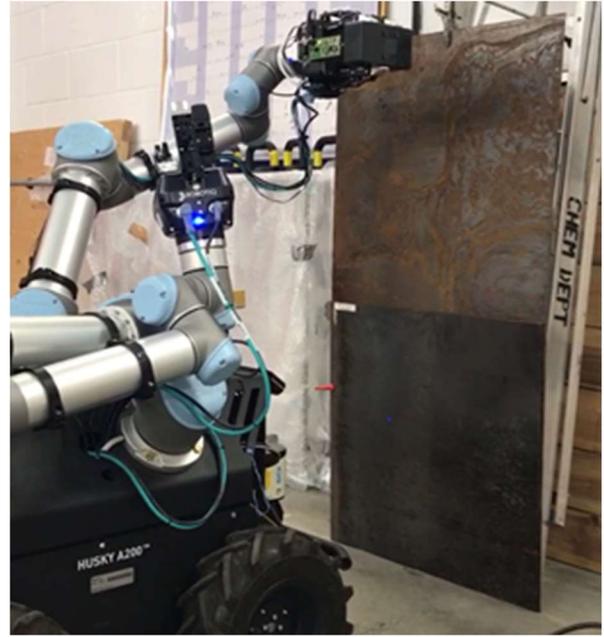

Figure 19 Husky A200 with a pair of UR5 manipulator arms integrated with the FMCW inspection tool during a corrosion inspection; the top sheet is lightly corroded steel and the lower sheet is non corroded and polished.

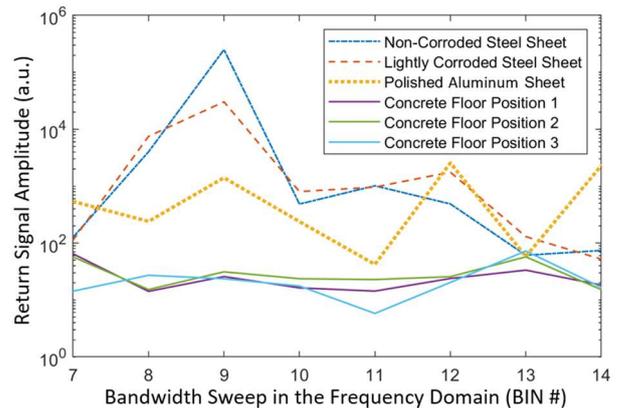

Figure 20 Observed return signal amplitude responses for differing metal and concrete targets. The peak amplitudes at BIN 9 represent the millimeter-wave response to targets at 10 cm from the FMCW radar sensor.

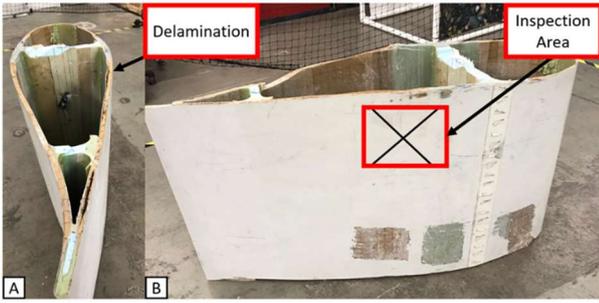

**Figure 21** A- Side on view of the decommissioned wind turbine blade displaying the delamination on the interior of the blade. B- Straight on view of the exterior of the decommissioned wind turbine blade highlighting the inspection area.

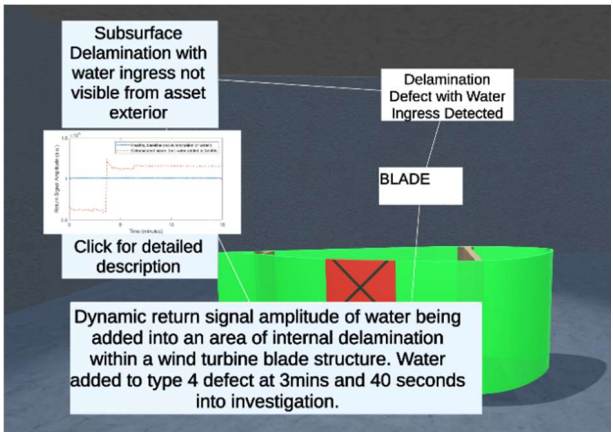

**Figure 22** AID indicating DT of acquired sandwich composite wind turbine blade in green highlighting a defective area of the blade with red hatching and options for a human operator to attain more information about the diagnosed fault.

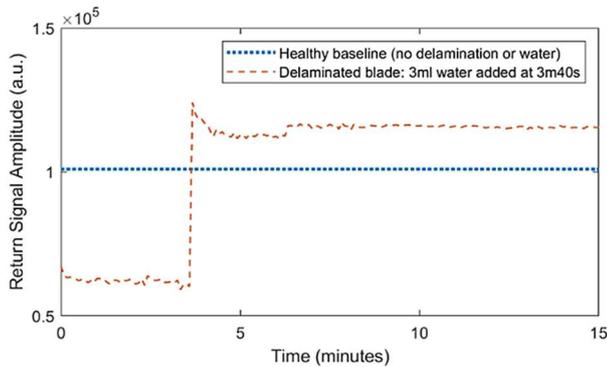

**Figure 23** Dynamic return signal amplitude of water ingress to an area of internal delamination within a wind turbine blade structure. Water added to type 4 defect at 3 mins and 40.

This application demonstrates an enhancement of $C^3$ governance, created through the transfer of data collected from the inspection device into the AID post processing tool for improved human-robot interaction capabilities. This allows wind farm operators to view the data collected in the synthetic environment of the real asset, achieving an improved operational overview, and leading to easy identification and localization of faults on wind turbine blades.

The application of millimeter wave sensing to offshore asset integrity represents a unique showcase of the capabilities of deployed, edge analysis sensors and their role in the wider, robotically deployed integration of data to the synthetic environment.

**VI. CONCLUSION**

Our review of the state-of-the-art in RAS shows that the predominant mode of robotic deployment for offshore wind farms utilizes COTS robotic systems. While this results in short developmental sprints and rapid deployment of customized RAS, we identify that the use of COTS systems limit deployed robotics in the offshore sector to simple, short term and typically individual roles.

As a result of our extensive review from academic and industrial sources into offshore robotics, we identify several key barriers to enhance semi and fully autonomous capabilities. These are run-time safety compliance, resilience and reliability. Due to the dynamic environment, the need for adaptive mission planning and intrinsic time variable reliability of resident robotic systems, a CPS approach based on our SSOSA is designed, implemented and verified to address these challenges.

$C^3$ limitations lead us to the creation of our novel SSOSA. We define our SSOSA, to address the following barriers imposed by limitations in $C^3$, these are: functional, operational, safety and planning requirements, which ensure mission resilience and self-certification. Via the design and implementation of a dynamic mission evaluation, we verify that our system of systems approach addresses safety, reliability, productivity and provides a run-time operational assessment of an offshore wind farm analogue. Thus, our SSOSA represents a new methodology that aggregates autonomous platforms, sensing, reliability modelling and HRI into one CPS. This amalgamates previously partitioned sub-elements into a common and synchronized DT environment. Our SDA incorporates the outputs from up to 1000 individual sensors and systems to create a hyper-enabled capability, enhancing visibility and increasing the ability of the autonomous system to query its operating environment and adapt its response accordingly. As proven in its implementation, our results have verified the SSOSA and the ability to provide accurate mission state of health, mission status and foresight modelling capabilities. Crucially, our framework ensures safety during the transition from semi to fully autonomous robotics, via consistent adherence to our three capability criteria of RAS. Our DT enables increased human interaction and has intrinsic value due to its flexibility and scalability, allowing for platform agnostic integration with COTS robotic systems.

Our novel SDA advances resilient robotics and their future servitization for offshore wind and sector-wide commercialization. We also facilitate integrated action-gain trustworthiness by converting raw data captured from all connected devices, including RAS and other associated infrastructure elements, to actionable information. This is demonstrated in our autonomous mission evaluation to respond to warnings and faults induced on the robot. We utilized non-destructive FMCW radar sensing for asset integrity inspection, delivering an increased operational overview via our AID and providing intuitive representations of data for the offshore wind farm operator.

Our scalable, tunable and platform agnostic SDA can be further applied in our roadmap from *'Adapt and Survive'* to an *'Adapt and Thrive'* paradigm, where our research will focus on the design and development of distributed intelligence to explore new optimizations that result from operational disruption in the stochastic offshore environment. We will advance more complex scenarios resulting in the processing of more prescriptive solutions to unforeseen challenges aimed at the offshore wind sector. This will include optimization of data needs versus probabilistic risk.

To advance semi and fully autonomous operations we need to improve safety, reliability and resilience. Our SDA framework has been evidenced to show improvements in resilience of our autonomous mission evaluation as defined by our three capability statements. We show that further design of our SDA has the potential to address operational barriers in the deployment of trusted resident robots.

# APPENDIXES

APPENDIX I
EIGHT IDENTIFIED STAGES FOR THE CONFINED SPACE ASSET INTEGRITY INSPECTION

| | Mission Event | | Challenges | | | |
|---|---|---|---|---|---|---|
| Stage | Objective | Description | Functional | Operational | Planning | Safety |
| A | Pre-mission planning | Inspection area recon | Human operation of robotic platform to map op-area | Remote control operation onsite | Access required for human and robot | Onsite safety of human and robot |
| | | Confined space asset integrity mission | Operator positions waypoints on map with tasks to complete | Human interaction with DT to create mission | Operator requires good knowledge of plant and mission to create mission | Is the robotic platform suitable for the environment? |
| B | Mission start at base point | System idle awaiting orders | Wireless connectivity between DT and robotic platform to receive orders from operator | Reliable wireless communications | Basepoint approved as safe | Self-certification from robot that it is fully deployable |
| C | Transit to asset integrity scan 1 | Navigation to Asset inspection 1 waypoint | Navigation and mapping | Husky computes most efficient route to complete mission | Accessible waypoints selected by human operator | Safety compliance with environment, humans and infrastructure |
| D | Perform asset integrity scan 1 | Asset inspection 1 | Manipulator positioned for FMCW radar asset inspection | Requires sufficient clearance for maneuver | Direction of scan input by the human operator within the DT | Safe distance from infrastructure adhered to |
| E | Transit to integrity scan 2 | Confined space operation section | Proximity detectors warn of collision risk with structure | Continuous navigation through confined area | Requires platform to plan optimal route | Increased risk of navigational error or collision with infrastructure |
| F | Perform asset integrity scan 2 | Asset inspection 2 | Manipulator positioned for FMCW radar asset inspection | Requires sufficient clearance for maneuver | Direction of scan input by the human operator within the DT | Safe distance from infrastructure adhered to |
| G | Return transit to base point | Warning 1: Motor is hot | Motor in danger of failure | Reduced mobility Decreased power | Reprioritize mission objectives | Increased risk of loss, mission incompletion or collision |
| | | Warning 2: Software is consuming resources | Managing the limited computing resource | Drain on computational efficiency (processing power) | Ontology decision-making whether to reprioritize mission | Robotic platform in danger of loss or stranding |

| | | Warning 3: Low battery | Reduced current available for systems | Limited time to complete mission | Robotic platform removes objectives from mission plan | Robotic platform in danger of loss or stranding, incomplete mission |
|---|---|---|---|---|---|---|
| H | Mission End | Scenario 1: No warnings detected by reliability ontology - Mission success | Real-time bidirectional communication, Synchronization with DT | Robotic platform completes mission and returns to base point. Updates operator of successful mission | Robotic platform updates synthetic environment to apply acquired data to next mission plan | Ontology never detected any risks therefore risks are minimal |
| | | Scenario 2: Some warnings detected - Mission success | Real-time bidirectional communication, Synchronization with DT to allow human-in-the-loop to advise/overlook decision-making | Robotic platform completes mission and returns to base, Updates operator of successful mission and warnings to be considered | Ontology decision-making must be set to continue under the severity of those warning conditions | Integrity of the robotic platform could be compromised but mission still achievable |
| | | Scenario 3: Many warnings detected by reliability ontology - Robotic platform stops to ensure integrity of asset is maintained- Mission Failure | Real-time bidirectional communication, Synchronization with DT to inform human-in-the-loop of faults and impose recovery of platform | Robotic platform stops at current position where warning occurs to prevent failure. Mission incomplete | Ontology decision-making must be set to failure under the severity of those warning conditions | Integrity of the robotic platform is compromised and unable to complete the mission. Platform is required to be recovered by human or another platform |

APPENDIX II
WRITTEN ALGORITHMS FOR FAULT AND WARNING DETECTION

**Algorithm 1** Motor Temperature Fault Check
**Require:** Motor temperate does not exceed maximum threshold
**Ensure:** Motor temperature stays within safe boundaries
1:   **if** *temperature > maximum threshold value* **then**
2:     **mission stop**
3:     **notify user of mission end**
4:   **end if**
5:   **if** *temperature > critical threshold value* **then**
6:     **query operator if the mission should stop**
7:     **if** *input = yes*
8:       **mission stop**
9:     **else**
10:       **mission continues**
11:     **end if**
12:   **end if**

**Algorithm 2** Motor Temperature Warning Check
**Require:** Notify user if the temperature enters a critical threshold
**Ensure:** Motor temperature stays within safe boundaries
1:   **if** *temperature > critical threshold value* **then**
2:     **query operator if the mission should stop**
3:     **if** *input = yes*
4:       **mission stop**
5:     **else**
6:       **mission continues**
7:     **end if**
8:   **end if**

**Algorithm 3** Software Resource Fault Check
**Require:** System resources of the robotic platform do not exceed maximum threshold
**Ensure:** System resource usage of the robotic platform stays within safe boundaries
1:   **if** *process RAM usage % > maximum RAM usage % threshold*
2:   *value*
3:   **or** *process CPU usage % > maximum CPU usage % threshold*
4:   *value* **then**
5:     **mission stop**
6:     **notify user of mission end**
7:   **end if**

**Algorithm 4** Software Resource Warning Check
**Require:** Notify user if system resource usage enters a critical threshold
**Ensure:** System resource usage of the robotic platform stays within safe boundaries
1:   **if** *process RAM usage % > critical RAM usage % threshold*
2:   *value*
3:   **or** *process CPU usage % > critical CPU usage % threshold*
4:   *value* **then**
5:     **query operator if the mission should stop**
6:     **if** *input = yes*
7:       **mission stop**
8:     **else**
9:       **mission continues**
10:     **end if**
11:   **end if**

**Algorithm 5** Low Battery Level Fault Check
**Require:** Battery level does not reduce below critical threshold whilst in mission
**Ensure:** Mission stop before battery is completely drained
1:   **if** *battery SoC < critical threshold value* **then**
2:     **mission stop**
3:     **notify user of mission end**
4:   **end if**

**Algorithm 6** Low Battery Level Warning Check
**Require:** Warn user of battery level entering warning threshold
**Ensure:** Mission stop before battery is completely drained
1:   **if** *battery SoC < warning threshold value* **then**
2:     **query operator if the mission should stop**
3:     **if** *input = yes*
4:       **mission stop**
5:     **else**
6:       **mission continues**
7:     **end if**
8:   **end if**

## ACKNOWLEDGEMENTS

This research was funded by the Offshore Robotics for the Certification of Assets (ORCA) Hub [EP/R026173/1], EPSRC Holistic Operation and Maintenance for Energy (HOME) from Offshore Wind Farms and supported by MicroSense Technologies Ltd (MTL) in the provision of their patented microwave FMCW sensing technology (PCT/GB2017/053275) and decommissioned wind turbine blade section.